\documentclass{article}

\PassOptionsToPackage{square,numbers}{natbib}

\usepackage[preprint]{neurips_2022}

\usepackage[utf8]{inputenc} 
\usepackage[T1]{fontenc}    
\usepackage{hyperref}       
\usepackage{url}            
\usepackage{booktabs}       
\usepackage{amsfonts}       
\usepackage{nicefrac}       
\usepackage{microtype}      
\usepackage{xcolor}         
\usepackage{amsmath}
\usepackage{adjustbox}
\usepackage{subfig}
\usepackage{diagbox}
\usepackage{todonotes}

\usepackage{bm}
\usepackage{float}
\usepackage{graphicx}

\newcommand{\ie}{\textit{i.e.}, }
\newcommand{\eg}{\textit{e.g.}, }

\newcommand{\ql}[1]{\textcolor{black}{#1}}
\newcommand{\ye}[1]{\textcolor{black}{#1}}

\newcommand{\xm}[1]{\textcolor{black}{#1}}
\newcommand{\revtwo}[1]{\textcolor{black}{#1}}
\begin{document}

\title{Deep Auto-encoder with Neural Response}
\author{
Xuming Ran$^1$,
Jie Zhang$^{2,3}$,
Ziyuan Ye$^{1}$,
Haiyan Wu$^{4}$,
Qi Xu$^{5}$, \\
\textbf{Huihui Zhou}$^{6,\ast}$,
\textbf{Quanying Liu}$^{1,\ast}$
}
\maketitle

\begin{centering}
$^1$Southern University of Science and Technology, China\\
$^2$Shenzhen Institutes of Advanced Technology, Chinese Academy of Sciences, China\\
$^3$University of the Chinese Academy of Sciences, China\\
$^4$Centre for Cognitive and Brain Sciences and Department of Psychology, University of Macau\\
$^5$School of Artificial Intelligence, Dalian University of Technology, China\\
$^6$Pengcheng Laboratory, China\\
$^\ast$Corresponding author:\texttt{liuqy@sustech.edu.cn to} Q.L. and \texttt{ zhouhh@pcl.ac.cn} to H.Z.
\end{centering}

\begin{abstract}
Artificial neural network (ANN) is a versatile tool \ql{to study the neural representation} in the ventral visual stream, \ql{and the knowledge in neuroscience in return inspires ANN models to improve performance in the task}. 
However, \ql{it is still unclear} how to merge these two directions into a unified \revtwo{framework}. 
In this study, we propose an integrated framework called Deep Autoencoder with Neural Response (DAE-NR), which incorporates information from ANN and the visual cortex to achieve better image reconstruction performance and higher neural representation similarity between biological and artificial neurons.
The same visual stimuli (\ie natural images) are input to both the mice brain and DAE-NR. 
The encoder of DAE-NR jointly learns the dependencies from neural spike encoding and image reconstruction. For the neural spike encoding task, the features derived from a specific hidden layer of the encoder are transformed by a mapping function to predict the ground-truth neural response under the constraint of image reconstruction. Simultaneously, for the image reconstruction task, the latent representation obtained by the encoder is assigned to a decoder to restore the original image \xm{under the guidance of neural information}. In DAE-NR, the learning process of encoder, mapping function and decoder are all implicitly constrained by these two tasks. 
Our experiments demonstrate that \textit{if and only if} with the joint learning, DAE-NRs can improve the performance of \xm{visual} image reconstruction and increase the \ql{representation} similarity between biological neurons and artificial neurons. 
The DAE-NR offers a new perspective on the integration of computer vision and neuroscience.
\end{abstract}

\section{Introduction}
Computer vision has achieved almost comparable performance to the human visual system on some tasks, mainly thanks to recent advances in deep learning.
Image reconstruction is one of the essential tasks in computer vision~\cite{wang2018image,Ravishankar2020ImageRF}. 
As a solution, the auto-encoder (AE) framework embeds the high-dimensional input to a low-dimensional latent space by the encoder and then reconstructs the image by the decoder~\cite{hinton2006reducing,goodfellow2016deep}.
Inspired by neuroscience, computer vision researchers have been interested in how to use information from biological neurons to achieve brain-like performance (such as robustness and ability to learn from small samples). The biology-inspired AE models may help improve performance in image reconstruction tasks and bring biological interpretability~\cite{Federer2020ImprovedOR,Schrimpf2018BrainScoreWA,safarani2021towards}. To this end, the key question is how to integrate biological information into AEs.

On the other hand, computational neuroscience is interested in building models that map stimuli to neural responses. Traditional models have difficulty expressing nonlinear characteristics between stimulus and neural response. Deep learning empowers computational neuroscience models and reveals the relationship between stimuli and neural spikes~\cite{klindt2017neural} 
Although biological and artificial neural networks may have fundamental differences in computation and learning~\cite{macpherson2021natural}, both are realized by interconnected neurons: the former by biological neurons; the latter by artificial neurons. Some previous work has focused on investigating the similarities between the information representation of biological neurons and artificial neurons using end-to-end ANNs, suggesting that artificial neurons in different layers of ANNs share similar representations with biological neurons in brain regions along ventral visual pathway~\cite{DiCarlo2012HowDT,Yamins2016UsingGD,walker2019inception,bashivan2019neural}. However, how to build artificial neural networks most similar to biological neural representations remains an open question.

Some studies have leveraged real neural responses to speed up the training process and improve network performance in object detection tasks~\cite{Federer2020ImprovedOR,Schrimpf2018BrainScoreWA,safarani2021towards}. However, no model has yet utilized neural responses as constraints to improve image reconstruction performance. 
More importantly, to our knowledge, there is no unified framework that can leverage both neural responses to improve model performance and image reconstruction tasks to improve representational similarity between biological and artificial neurons.

\begin{figure*}[t]
    \centering
    \includegraphics[width=0.99\textwidth]{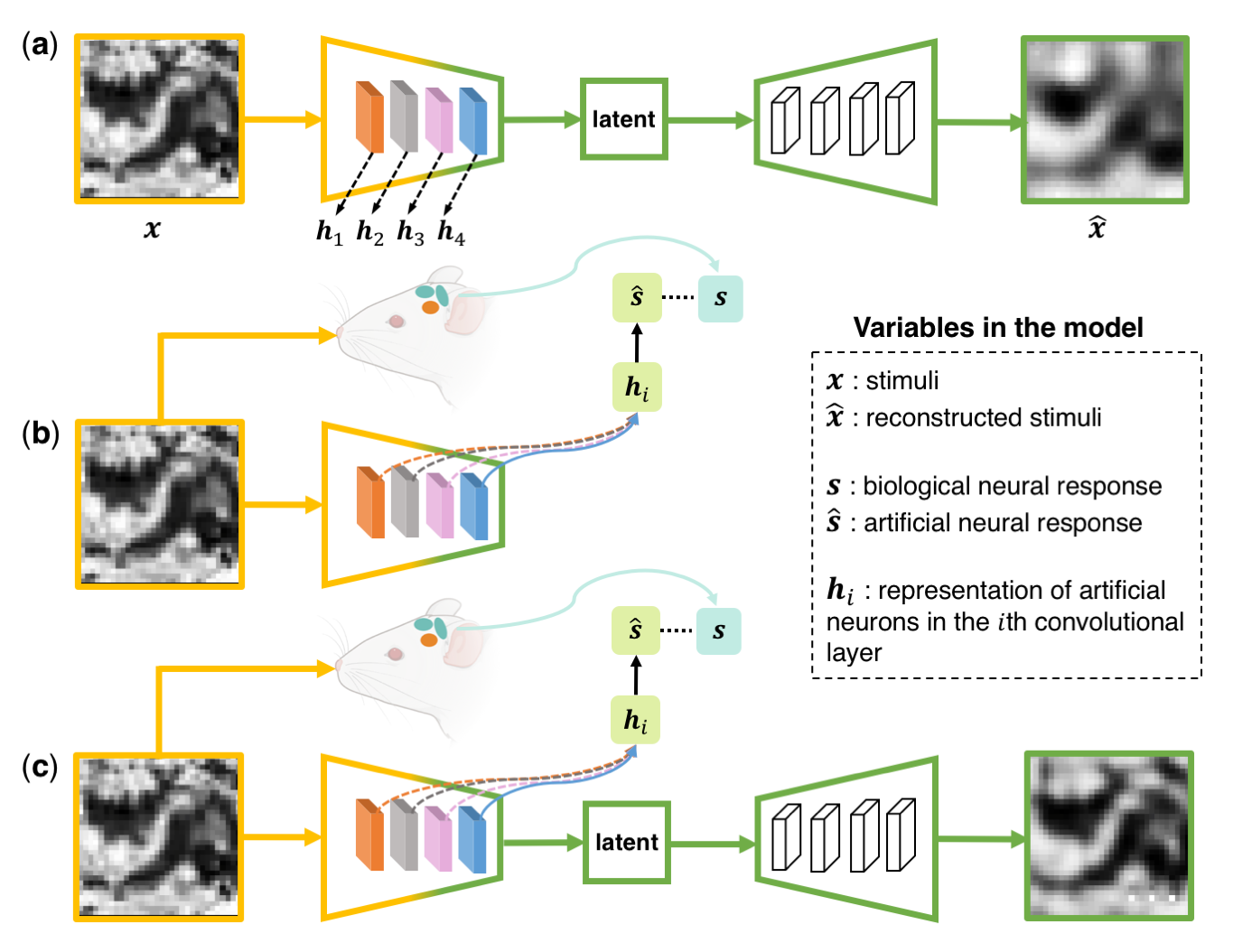}
    \caption{The illustration of the model of (a) the standard deep auto-encoder (DAE) for images reconstruction; (b) the convolutional neural network with factorized readout (CNN-FR) for prediction of neuron responses; (c) the DAE with the neuron response (DAE-NR) for images reconstruction and predictions of neuron responses. $\bm{s}$ is the biological neural response, the prediction of biological neural response is represented as $\bm{\hat{s}}$, and $\textbf{h}_{i}$ ($i \in \{1, 2, 3, 4 \}$) is the feature of the $i$th convolutional layer.}
    \label{fig:my_model}
\end{figure*}
In this paper, we aim to tackle these two questions in one piece. Specifically, we propose a united biologically inspired framework, which jointly learns i) to project features \xm{of visual input} in a specific layer of the encoder to biological neural responses by a mapping function and ii) to reconstruct the visual input via the decoder. As a result, its encoder has a higher representational similarity to the real neural responses, and its decoder achieves better image reconstruction performance.
Our contributions can be summarized as follows.
\begin{itemize}
    \item We present a \revtwo{biologically inspired} framework called Deep Auto-encoder with Neural Response (DAE-NR). The \revtwo{framework} can simultaneously learn to predict neural responses and to reconstruct the visual stimuli (\textbf{Sec.\ref{sec:method}}).
    \item \ql{The deep auto-encoders embedded in DAE-NR can improve the image reconstruction quality with the help of a Poisson loss on the predicted neural activity, compared to the baseline auto-encoder models} (\textbf{Sec.\ref{sec:results_imager_construction} \& \ref{sec:results_imager_construction_resneu}}).
    \item \ql{The computational neuroscience model (CNM) via DAE-NR offers higher resemblance between artificial neurons and biological neurons compared to the competing end-to-end CNM models} (\textbf{Sec.\ref{sec:results_neural_similarity} \& \ref{sec:results_imager_construction_resneu}}).
\end{itemize}

\section{Related work}




\textbf{Image reconstruction via auto-encoders:} 
A big breakthrough in image reconstruction is in ~\cite{hinton2006reducing}, which equips an autoencoder with a stack of restricted Boltzmann machines.
The denoising autoencoder~\cite{vincent2008extracting} and convolutional autoencoder \xm{(CAE)}~\cite{masci2011stacked} further improve the stability of autoencoders by adding noise and taking advantage of convolutional layers in feature extraction, respectively. 
Variational Autoencoder (VAE) enhances model robustness of generating insightful representations through a latent distribution learning mechanism~\cite{Kingma2014AutoEncodingVB}. 
Vector quantisation VAE (VQ-VAE) employs the vector quantisation to obtain a discrete latent representation that can improve the quality of image reconstruction and generation~\cite{Oord2017NeuralD}.
Recent advances follow a similar trend of latent space exploration and generalize variants of DAE to many domains~\cite{larsen2016autoencoding,khattar2019mvae,park2020swapping,wang2021encoder,cai2021unified,ran2021detecting}.
However, However, DAE and its variants in image reconstruction suffer from the same problem; the parameters have a high degree of freedom.
In other words, parameters can only be learned through error backpropagation guided by gradients.
Therefore, some meaningful constraints on the parameter space will be beneficial for learning.

\begin{figure*}[t]
    \centering
    \includegraphics[width=0.9\textwidth]{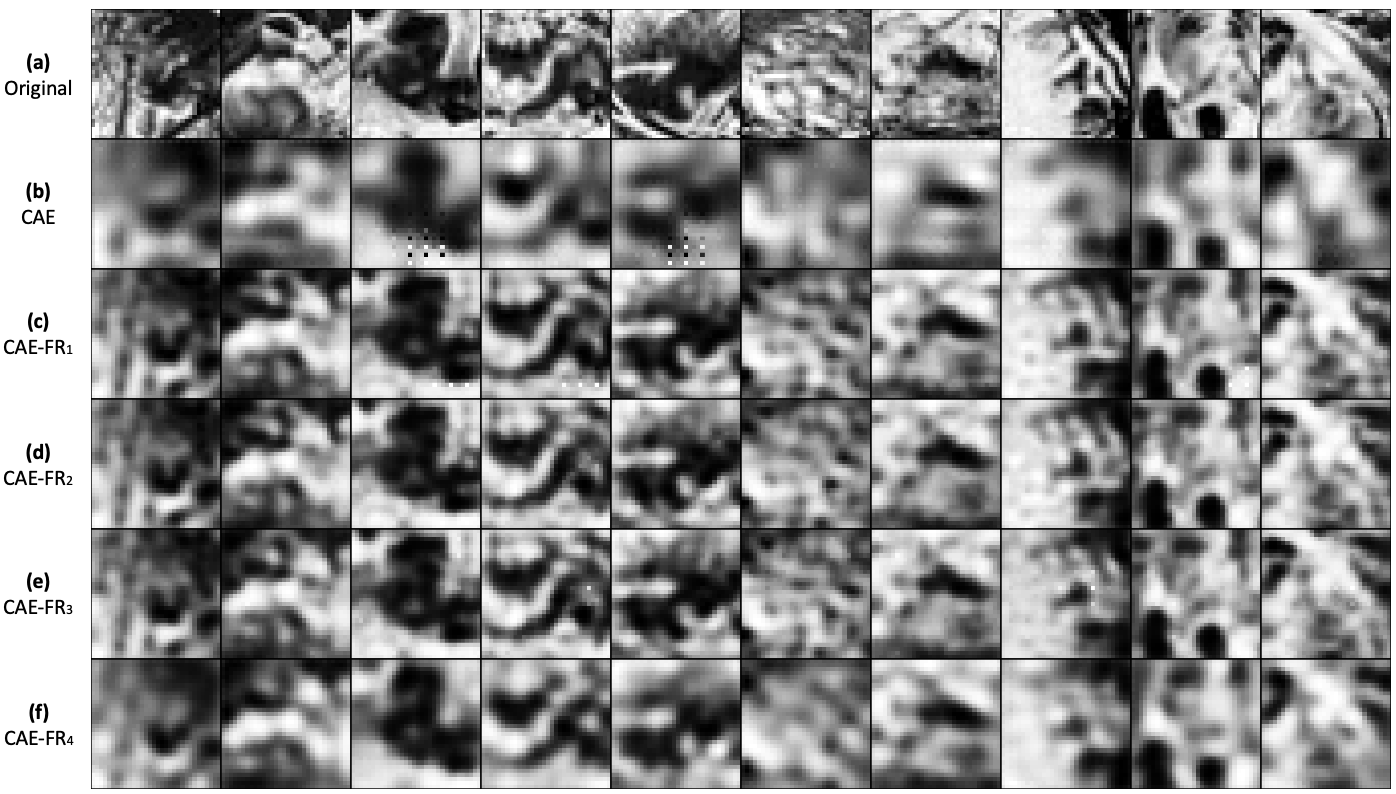}
    \caption{The reconstructed images with neurons in Region 3. From top to bottom, each row displays (a) the original images, the images reconstructed by (b) CAE, (c) CAE-FR$_{1}$, (d) CAE-FR$_{2}$, (e) CAE-FR$_{3}$, (f) CAE-FR$_{4}$, respectively.}
    \label{fig:imgrecreg3}
\end{figure*}

\textbf{Neural similarity \ql{in computational neuroscience}:}
Many models in computational neuroscience have been proposed to build a relationship between stimuli and the corresponding neural spike responses, which are known as neural spike encoding. The goal of these models is to increase the similarity between predicted and true neural responses. Historically, much effort has been devoted to finding tuning curves for specific features of visual stimuli, such as the orientation of bars, to predict neural responses~\citep{Carandini2005DoWK,Drger1975ReceptiveFO,hubel1962receptive,Hubel1968ReceptiveFA,Niell2008HighlySR}. 
This is feasible for some neurons in the primary visual cortex, but not for all neurons.
Nonlinear methods provide a more general approach to predicting neural responses, including energy models~\cite{hubel1962receptive}, the linear-nonlinear (LN) model and its extended version LN-LN model~\cite{meyer2017models}. Traditional machine learning methods, such as generalized linear models (GLMs)~\citep{Willmore2008TheBW}, the multi-layer perceptron (MLP) and the support vector regression (SVR)~\cite{das2019computational}, have been brought into the computational neuroscience field to improve the neural similarity. 
More recently, a similar hierarchical structure has been found in ventral visual pathway~\citep{DiCarlo2012HowDT,Vintch2015ACS,Rowekamp2017CrossorientationSI} and in deep convolutional neural networks~\citep{Fukushima1983NeocognitronAN,LeCun1989HandwrittenDR,Krizhevsky2012ImageNetCW}. 
The brain and CNNs may share similar neural representations for feature extraction from stimuli, layer by layer, from simple to complex.  Inspired by this, Yamins el al.\cite{Yamins2016UsingGD} proposed to use hierarchical CNNs as computational neuroscience models to investigate the neural similarity. They reported neural representation similarity between the biological neurons along the ventral visual stream and biological neurons in different convolutional layers~\cite{Yamins2016UsingGD}. 
CNN with a fully-connected readout layer (CNN-FC) can map the convolutional features to neural responses~\cite{mcintosh2016deep}. However, such a fully-connected readout layer typically contains a large number of parameters.
To reduce the number of parameters in the readout layer, the CNN with a factorized readout layer (CNN-FR) factorizes the convolutional feature into a spatial mask~\cite{klindt2017neural}. The CNN with a fixed mask (CNN-FM) is a variant of CNN-FR. The computational neuroscience models based on convolutional and readout layers become an important tool for studying the neural similarity between the brain and CNN.


\section{Method}
\label{sec:method}

Here we first introduce the notation of data spaces and variables. We then present the framework of deep auto-encoders with the neural response (DAE-NR) in Sec.~\ref{subsec:dae-nr}. Finally, we describe realizations of the DAE-NR framework using CAE and CNN-FR in Sec.~\ref{subsec:cae-fr}.

The data space of visual stimuli, the neural responses, and the features of stimuli in the $i$-th convolutional layer are denoted by $\mathcal{X}$, $\mathcal{S}$, and $\mathcal{H}_{i}$. The visual stimuli (\ie natural images) and the corresponding neural responses (\ie neural spikes in the V1 region) are represented as $\bm{x} \in \mathbb{R} ^{N \times P \times P \times C } $ and $\bm{s} \in \mathbb{R} ^{N \times M } $, respectively. 
The features of stimuli in the $i$-th convolutional layer are denoted as $\bm{h}_{i} \in \mathbb{R}^{N \times K \times K \times F } $, with the sample size ($N$), the image resolution ($P$), the image channel ($C$), the kernel size of a feature of stimuli ($K$), the number of kernels ($F$), and the number of V1 neurons ($M$), respectively.

\subsection{The DAE-NR framework}
\label{subsec:dae-nr}

The DAE-NR combines the function of DAE for image reconstruction and the role of CNM for predicting neural responses. The DAE-NR framework consists of three parts, including \ql{an encoder} $f_1$: $\mathcal{X} \rightarrow \mathcal{H}_i$,  \ql{a decoder} $f_2$: $\mathcal{H}_i \rightarrow \mathcal{\hat{X}}$, and \ql{a mapping function} $f_3:\mathcal{H}_i \rightarrow \mathcal{\hat{S}}$.

\textbf{DAE:} The standard DAEs and their variants (\eg CAE~\cite{masci2011stacked}, VAE~\citep{Kingma2014AutoEncodingVB} and VQ-VAE~\citep{Oord2017NeuralD}) consist of an encoder, the latent space, and a decoder, which have become mainstream models for image reconstruction. In our work, instead of separating the encoder and decoder by the latent layer, we split the encoder and decoder at the $i$-th layer of DAE. In this way, the architecture of our DAE turns to be: (i) an encoder $f_1:\mathcal{X} \rightarrow \mathcal{H}_i$, to embed the input to neural representation in $i$-th layer; (ii) a decoder $f_2:\mathcal{H}_i \rightarrow \mathcal{\hat{X}}$, to reconstruct the input based on neural representations in $i$-th layer. We formally describe them as follows:
\begin{equation}
\begin{aligned}
	\ql{\text{the encoder: }} \ & \bm{h}_{i}=f_{1}(\phi_{\mathrm{1}},\bm{x}),\\
	\ql{\text{the decoder: }} \ &\hat{\bm{x}}_{}=f_{2}(\phi_{\mathrm{2}},\bm{h}_{i}),
\end{aligned}
\end{equation}
where the $\hat{\bm{x}}$ is reconstructed from the original image $\bm{x}$. Both the encoder and decoder are realized by neural networks with parameters $\phi_{1}$ and $\phi_{2}$, respectively. The goal of DAE is to reconstruct the image by optimizing the the loss function $\mathcal{L}(\phi_{1},\phi_{2})$.

\textbf{CNM:} The computational neuroscience models (\eg CNN-FR~\citep{klindt2017neural,bashivan2019neural}, CNN-FC~\citep{mcintosh2016deep}, and CNN-FM~\citep{klindt2017neural}) usually consist of the encoder and the readout layer. The encoder can be used to extract the image's features. The readout layer is used to map the feature space $\mathcal{H}_i$ to the neural responses in the space $\mathcal{S}$. The mapping function $f_3:\mathcal{H}_i \rightarrow \mathcal{\hat{S}}$ is 
\begin{equation}
\hat{\bm{s}}=f_3(\bm{h}_i, \theta ).
\end{equation}

We employ the $\mathcal{L}_{}(\theta)$ loss to optimize the \ql{representation} similarity between artificial \ql{neurons} (\ie $\hat{\bm{s}}$) and biological neurons \ql{neurons} (\ie $\bm{s}$) in the mapping function.

The loss function of DAE-NR \ql{explicitly considers \ql{both} the image reconstruction task in computer vision and the neural representation similarity task in computational neuroscience, as defined in Eq.~\eqref{eq:loss-DAE-NR}.}
\begin{equation}
\label{eq:loss-DAE-NR}
\mathcal{L}_{}(\phi_{1},\phi_{2},\theta)= \alpha *\mathcal{L}_{}(\phi_{1},\phi_{2})+ \beta*\mathcal{L}_{}(\theta), 
\end{equation}
where $\alpha$ and $\beta$ are the hyperparameters to trade off the image reconstruction task and the neural representation similarity task.  

\subsection{Realizations of DAE-NR}
\label{subsec:cae-fr}
Here, we first realize DAE-NR by a toy model (\ie CAE-FR) combining a convolutional autoencoder (CAE) and a CNN with factorized readout (CNN-FR). It is important to note that the DAE-NR framework is compatible with different instantiations. We have full implementation in Table~\ref{dae-nevariants}.

\textbf{CAE:} Both the \ql{encoder and the decoder} are realized by \ql{convolutional} neural networks with parameters $\phi_{1}$ and $\phi_{2}$, respectively. The loss function of CAE is formulated as a $L_2$ norm:
\begin{equation}
\mathcal{L}_{}(\phi_{1},\phi_{2})= \left\|\bm{x} - \hat{\bm{x}}\right\|_{2}^{2} =
\left\|\bm{x} - f_{2}(\phi_{\mathrm{2}},f_{1}(\phi_{\mathrm{1}},\bm{x}))
\right\|_{2}^{2},
\end{equation}
where the $\hat{\bm{x}}$ is reconstructed from the \ql{original image} $\bm{x}$.

\textbf{CNN-FR:}
The CNN-FR consists of two parts, the convolutional layers \ql{as the encoder} and the factorized readout layer~\citep{klindt2017neural,bashivan2019neural,Cadena2019DeepCM,Zhuang2021UnsupervisedNN}. The convolutional layers convolve the image with a number of kernels followed by batch normalization, \ql{resulting in multiple} feature maps. The readout layer pools the output of the convolutional layer (\ie $\bm{h_i}$) by applying a sparse mask \ql{on} each neuron. \ql{Let us denote that} $\bm{h}_i$ lies in the feature space $\mathcal{H}_i$ and the neural responses in the space $\mathcal{S}$. The mapping function is 
\begin{equation}
\hat{\bm{s}}=f_3(\bm{h}_i, \theta_{\mathrm{s}},\theta_{\mathrm{d}} )=\left[\sum\left(\theta_{\mathrm{s}} \cdot \bm{h}_i\right)\right] * \theta_{\mathrm{d}}+{\bm{b}},
\end{equation}
where $\theta_{\mathrm{s}}$ is the spatial mask, $\theta_{\mathrm{d}}$ is the weights sum of all features $\bm{h}_i$, and $\bm{b}$ is the bias. We use the Poisson loss to optimize the \ql{representation} similarity between artificial \ql{neurons} (i.e. $\hat{\bm{s}}$) and biological neurons \ql{neurons} (i.e. $\bm{s}$) in the mapping function as Eq.(\ref{eq:poisson-loss}),
\begin{equation}
\label{eq:poisson-loss}
\mathcal{L}_{}(\theta_{\mathrm{s}},\theta_{\mathrm{d}}) = \sum\left(\hat{\bm{s}}- \bm{s} \log \hat{\bm{s}} \right).
\end{equation}

Previous studies have shown that the responses of V1 neurons to natural stimuli are sparse, and the activity of neural populations with higher sparseness exhibits greater discrimination against natural stimuli.~\citep{Vinje2000SparseCA,Weliky2003CodingON,Froudarakis2014PopulationCI,Yoshida2020NaturalIA}. Likewise,~\citep{Zhuang2017DeepLP} has reported that the resemblance between the representation of biological neurons and artificial neurons in higher convolutional layers exists only under the sparsity constraint on the CNN, regardless of any other factors (\eg, model structure, training algorithm, receptive field size, and property of training stimuli). 
In our study, the representational similarity of V1 neurons is brought to a specific layer of the CAE encoder (\textbf{$\bm{h}_{i}$}, \textbf{$i \in [1,2,3,4]$}) with the sparsity constraint for artificial neurons in this layer. 

\textbf{CAE-FR:} The CAE-FR combines the function of CAE and CNN-FR. The loss of CAE-FR \ql{explicitly considers \ql{both} the image reconstruction task and the neural representation similarity task, as defined in Eq.~\eqref{eq:loss-CAE-FR}},
\begin{equation}
\label{eq:loss-CAE-FR}
\mathcal{L}_{}(\phi_{1},\phi_{2},\theta_{\mathrm{s}},\theta_{\mathrm{d}})= \alpha *\left\|\bm{x} - f_{2}(\phi_{\mathrm{2}},f_{1}(\phi_{\mathrm{1}},\bm{x}))
\right\|_{2}^{2}+ \beta*\sum (f_3(\bm{h}_i, \theta_{\mathrm{s}},\theta_{\mathrm{d}} )- \bm{s} \log f_3(\bm{h}_i, \theta_{\mathrm{s}},\theta_{\mathrm{d}} )).
\end{equation}
Intuitively, \ql{the larger $\alpha$ favors the reconstruction task, while the larger $\beta$ biases toward the neural representation similarity task.}

\begin{table}[htbp]
\centering
\caption{The variants of DAE-NR. The implementation details of CAE-FR are presented in Sec ~\ref{subsec:cae-fr}. The implementation of other variants is the same as CAE-FR.}
\begin{tabular}{l|ccc}
\hline
\multicolumn{1}{l}{\diagbox{CNMs}{DAEs}} & \multicolumn{1}{|c}{CAE~\cite{masci2011stacked}} & \multicolumn{1}{c}{VAE~\citep{Kingma2014AutoEncodingVB}} & \multicolumn{1}{c}{VQ-VAE~\citep{Oord2017NeuralD}} \\ \hline

CNN with factorized readout~\citep{klindt2017neural} (CNN-FR)                                       & \textbf{CAE-FR}                  & VAE-FR                   & VQ-VAE-FR                   \\ 
CNN with fully-connected readout~\citep{mcintosh2016deep} (CNN-FC)                                       & CAE-FC                   & VAE-FC                   & VQ-VAE-FC                   \\ 
CNN with fixed mask~\citep{klindt2017neural} (CNN-FM)                                       & CAE-FM                   & VAE-FM                   & VQ-VAE-FM                   \\ \hline
\end{tabular}
\label{dae-nevariants}
\end{table}

\section{Experiments and Results}
\subsection{Experimental settings}
\textbf{Datasets:} We conduct experiments on a publicly available dataset with the gray-color images as visual stimuli and the corresponding neural responses. The neural response dataset is obtained from~\cite{Antolk2016ModelCB}. The neural responses in the dataset are recorded in three regions in the primary visual cortex (V1) of sedated mice visually stimulated with the natural images (See Appendix Fig.~\ref{fig:trnimg}). The number of neurons over the three brain regions is shown in the Appendix Table~\ref{tab:numregs}.

\textbf{Models:} We realized DAE-NRs with nine different combinations of DAE and CNM (See Table~\ref{dae-nevariants}). DAE-NRs are compared with DAE baselines for the image reconstructions task and compared with CNM baselines for the neural representation similarity task. We choose CAE, VAE and VQ-VAE as the DAE baselines and CNN-FR, CNN-FC and CNN-FM as the CNM baselines, respectively. 
Especially, we have the CNM with $\textbf{h}_{i}$ (CNM$_{i}$, $ i \in \{1,2,3, 4\}$). The CNM in DAE-NR can be readout from four different convolutional layers (\ie $\textbf{h}_{i}$ of the DAE encoder), resulting in four variants of DAE-NRs (\ie DAE-NR$_{i}$, $ i \in \{1,2,3, 4\}$) which represent the CNM extracted from the different convolutional layers $\textbf{h}_{i}$.

\textbf{Network architectures:} The architecture of the DAE in DAE-NR is 48C23-48C13-48C23-48C13-48DC13-48DC23-48DC13-1DC23 ($k$C${nm}$ and $k$DC${nm}$ indicates $k$ filters in the convolutional and deconvolutional layer, $n$  stride and $m$ filters). 
The dimension of latent variable in CAE and VAE is set as 100; the prior of VAE and VQ-VAE is the Gaussian distribution $N(0,1)$ and uniform distribution, respectively. The dimension of latent embedding space in VQ-VAE is $32\times256$.
Each convolutional layer is followed by a batch normalization with an ELU activation function, while the activation function in the final layer for image reconstruction is $tanh$. The CNM in DAE-NR shares the four different convolutional layers ($\textbf{h}_{1}$, $\textbf{h}_{2}$, $\textbf{h}_{3}$, and $\textbf{h}_{4}$). 
The CNM in DAE-NR can be readout from four different convolutional layers (i.e. $\textbf{h}_{i}$ of the encoder of DAE), so there are four variants of DAE-NR, i.e. DAE-NR$_{i}$, $ i \in \{1,2,3, 4\}$ representing the CNM  extracted from the different convolutional layers $\textbf{h}_{i}$.

\textbf{Training procedures:} We preprocess the images (i.e., reshape the size of the natural image to $32 \times 32 \times 1$ and normalize the intensity of the image to [-1,1]) and then input them to the model. \ql{The model is trained with} an initial learning rate of 0.001, and the early stopping \ql{strategy} is applied based on a separated validation set. If the error in the validation set does not improve by 1000 steps, we return the best parameter set, reduce the learning rate by two, and train in the second time.The VAE is optimized by the $ELBO$, while the VQ-VAE is optimized following the settings in~\citep{Oord2017NeuralD}. The settings of hyperparameters for tasks in  Sec~\ref{sec:results_imager_construction} and Sec~\ref{sec:results_neural_similarity}  are detailed in Appendix Table~\ref{tab:setexpimgrec} and \ref{tab:setexpneusim} , respectively. 
The Appendix  Table~\ref{tab:setexpimgrecdarvari3} and Table~\ref{tab:setexpneusimreg1}\&\ref{tab:setexpneusimreg2}\&\ref{tab:setexpneusimreg3} are the settings the hyperparameters of different DAE-NR variants for image reconstruction on Region 3 and  neural similarity
experiments on Region 1,2 and 3 in Sec~\ref{sec:results_imager_construction_resneu}, respectively.

\textbf{Tasks:} There are two tasks in the experiment: 1) the image reconstruction (IR) task and 2) the neural representation similarity (NRS) task.
We apply CAE-FR to analyse the effects of neural responses on IR and the effects of image reconstruction on NRS tasks in DAE-NR. We compare the other DAE-NR variants with DAE and CNM for IR and NRS tasks to explore the generalisation capability of DAE-NR.
In the  IR task, we use the mean squared error (MSE$\downarrow$), structural similarity (SSIM$\uparrow$)~\cite{Wang2004ImageQA}, and peak signal-to-noise ratio (PSNR$\uparrow$)~\cite{Wang2004ImageQA} \ql{as metrics} to quantify the image reconstruction performance\footnote{The up arrow $\uparrow$ indicates that \ql{the higher the value, the better, and the down arrow $\downarrow$ indicates that the lower, the better.}}. We compare the DAE-NR$_{1}$, DAE-NR$_{2}$, DAE-NR$_{3}$ and DAE-NR$_{4}$ with the standard DAE.
\ql{In the NRS task, we use} Pearson correlation coefficient (PCC$\uparrow$) as a metric to quantitatively evaluate models. \ql{We implement the traditional end-to-end CNM$_{i}$ ($i \in \{ 1, 2, 3, 4 \}$) as baseline models}.



\begin{table*}[htbp]
\centering
\caption{The quantitative results of image reconstruction with all neurons in the region 1, 2, and 3, respectively. The best result in each region under different metrics is highlighted with boldface.}
\begin{adjustbox}{width=\columnwidth}
\begin{tabular}{c|ccc|ccc|ccc}
\hline
 Region&
  \multicolumn{3}{c|}{Region 1} &
  \multicolumn{3}{c|}{Region 2} &
  \multicolumn{3}{c}{Region 3} \\ \hline
\diagbox{Model}{Metric} &
  MSE$\downarrow$ &
  PSNR$\uparrow$&
  SSIM$\uparrow$&
  MSE$\downarrow$ &
  PSNR$\uparrow$ &
  SSIM$\uparrow$&
  MSE$\downarrow$ &
  PSNR$\uparrow$ &
  SSIM$\uparrow$ \\ \hline
\multicolumn{1}{c|}{CAE} &
  \multicolumn{1}{c}{0.022} &
  \multicolumn{1}{c}{23.709} &
  \multicolumn{1}{c|}{0.771} &
  \multicolumn{1}{c}{0.024} &
  \multicolumn{1}{c}{23.338} &
  \multicolumn{1}{c|}{0.754} &
  \multicolumn{1}{c}{0.081} &
  \multicolumn{1}{c}{17.039} &
  \multicolumn{1}{c}{0.561} \\
CAE-FR$_{1}$ &
  \textbf{0.021} &
  \textbf{23.829} &
  \textbf{0.776} &
  \textbf{0.023} &
  23.392 &
  0.753 &
  0.044 &
  19.751 &
  \multicolumn{1}{l}{0.763} \\ 
CAE-FR$_{2}$ &
  \textbf{0.021} &
  23.779 &
  0.775 &
  \textbf{0.023} &
  23.440 &
  0.759 &
  \textbf{0.043} &
  \textbf{19.819} &
  \textbf{0.764} \\ 
CAE-FR$_{3}$ &
  \textbf{0.021} &
  23.778 &
  0.775 &
  0.024 &
  23.330 &
  0.755 &
  \textbf{0.043} &
  19.789 &
  0.761 \\
CAE-FR$_{4}$ &
  0.022 &
  23.721 &
  0.773 &
  \textbf{0.023} &
  \textbf{23.491} &
  \textbf{0.760} &
  0.059 &
  18.462 &
  0.668 \\ \hline
\end{tabular}
\end{adjustbox}
\label{tab:recon-results1}
\end{table*}

\begin{figure}[htbp]
\centering
\includegraphics[page=1,width=2.5in]{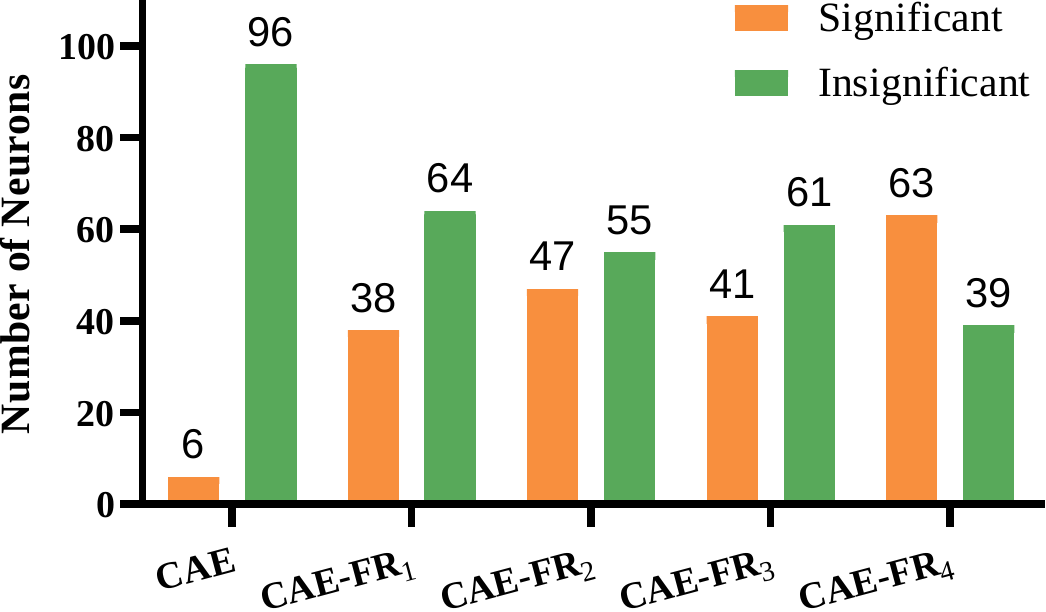}
\caption{The number of significant \ye{neurons} and insignificant neurons of region 3 in the image reconstruction task. \ye{The threshold for significance is $p\leq0.05$.}}
\label{fig:imgrecinsginificant}
\end{figure}

\begin{figure*}[htbp]
\centering
  \subfloat{\label{fig:subfig:1}
    \includegraphics[page=1,width=1.8in]{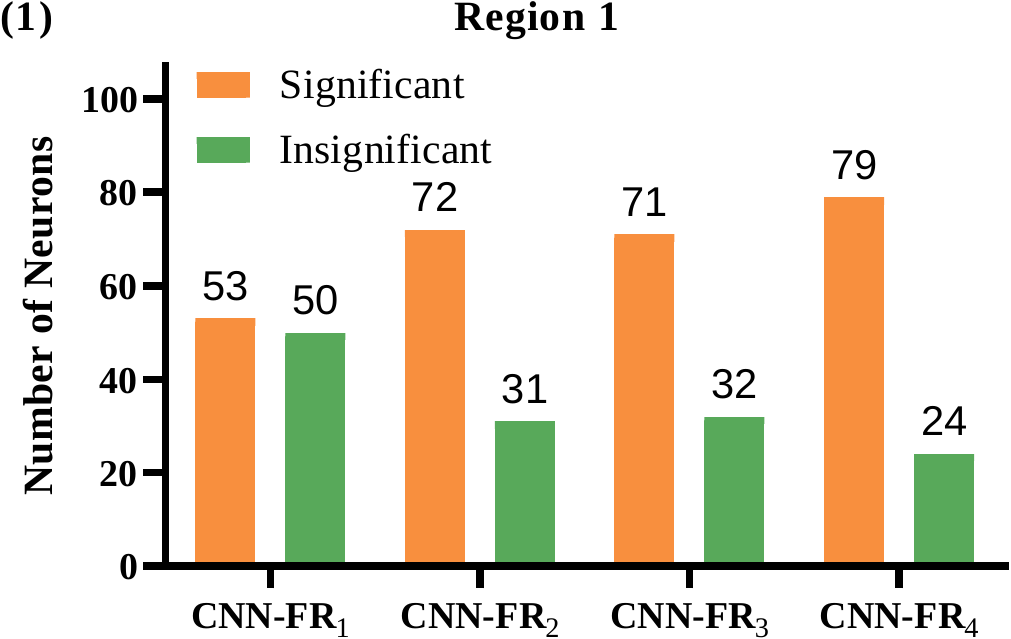}}
  \subfloat{\label{fig:subfig:2}
     \includegraphics[page=1,width=1.8in]{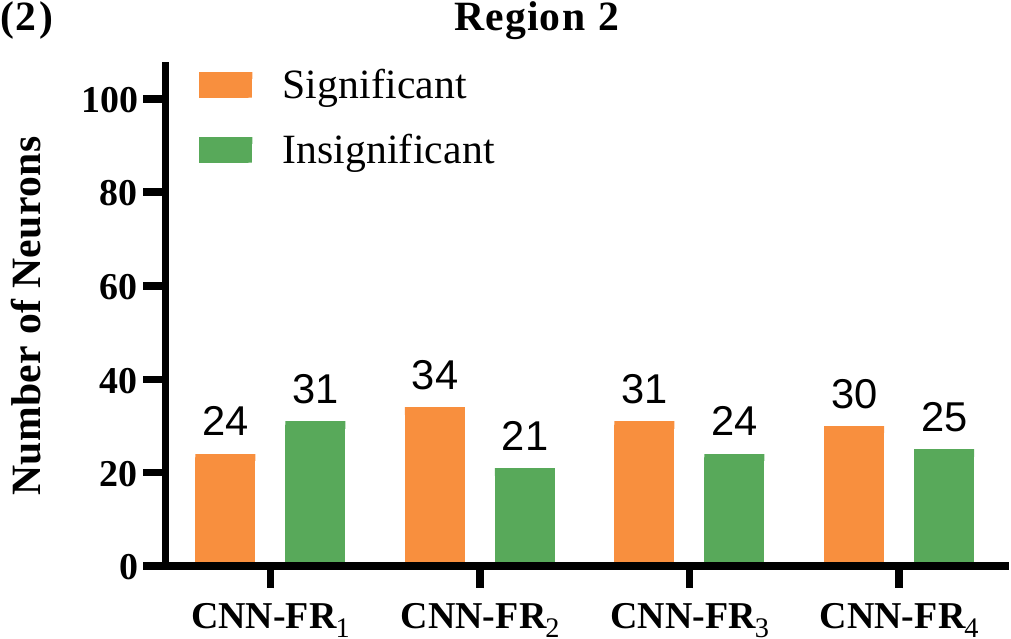}}
  \subfloat{\label{fig:subfig:3}
     \includegraphics[page=1,width=1.8in]{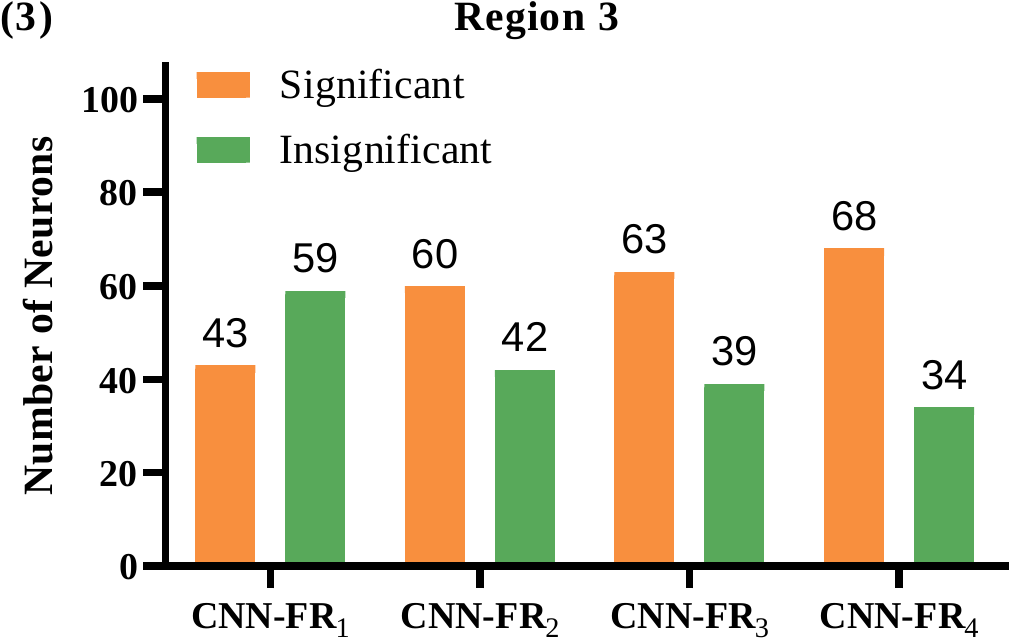}}
     
 \subfloat{\label{fig:subfig:4}
    \includegraphics[page=1,width=1.8in]{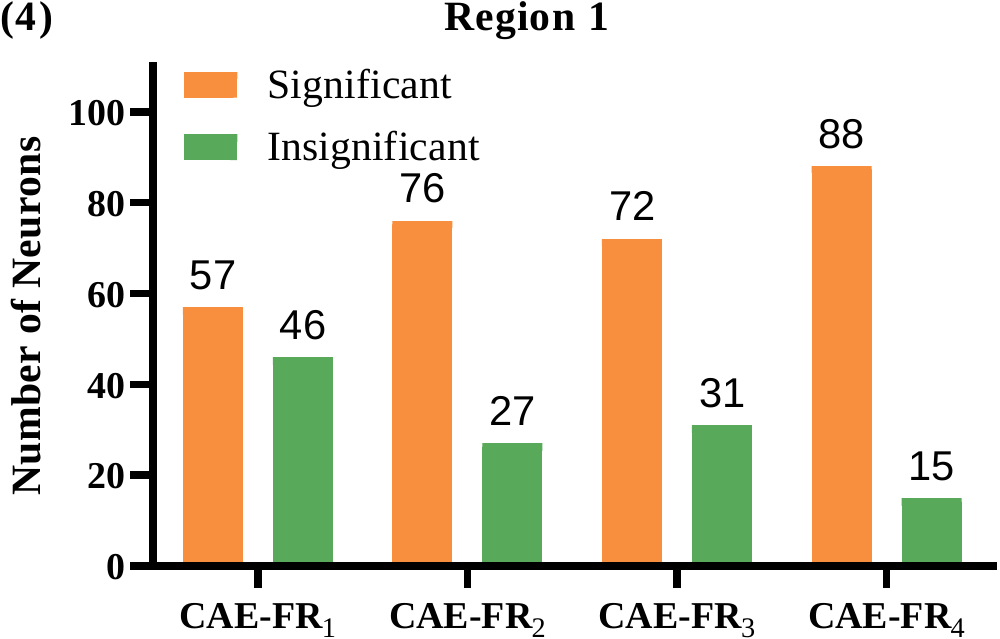}}
  \subfloat{\label{fig:subfig:5}
     \includegraphics[page=1,width=1.8in]{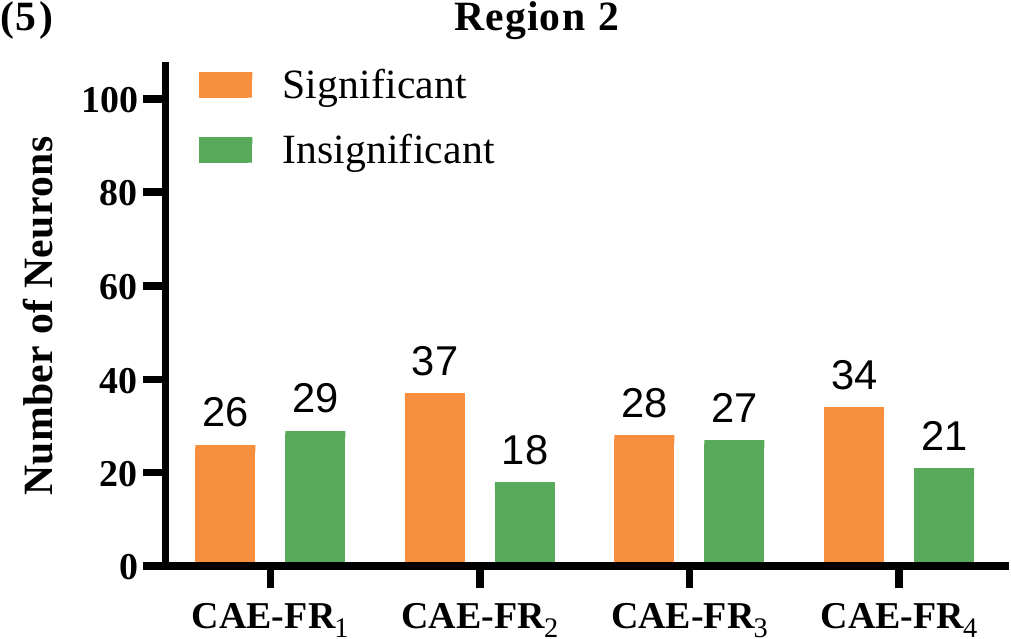}}
  \subfloat{\label{fig:subfig:6}
     \includegraphics[page=1,width=1.8in]{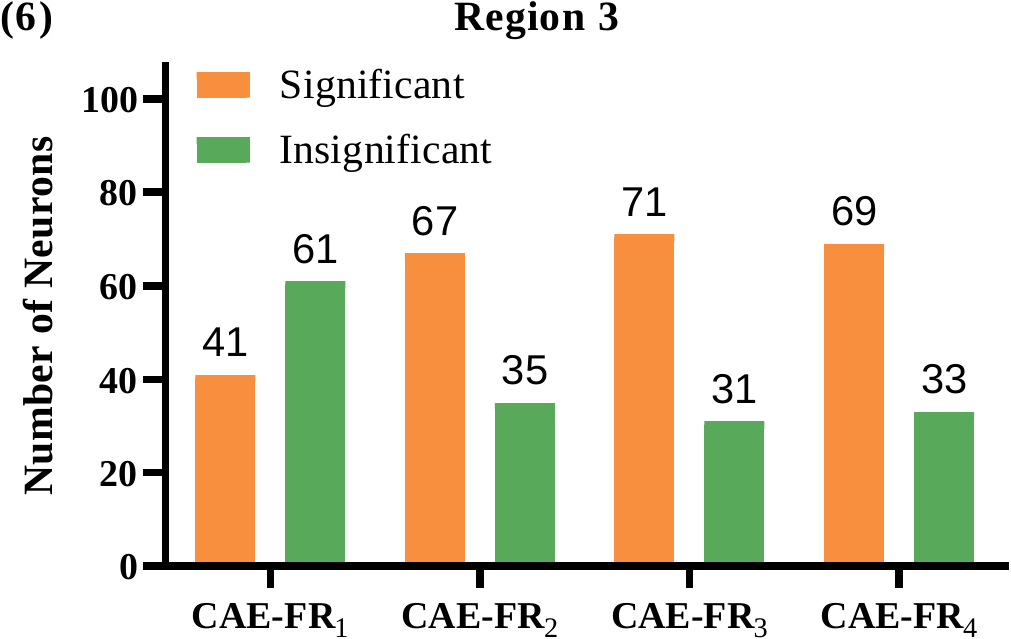}}
\caption{The number of significant \ql{neurons} (orange color) and insignificant  neurons (green color) in region 3 in the image reconstruction task. \ql{The threshold for significance is $p \leq 0.05$}. The first row are \ql{the} baseline models (CNN-FRs) and the second row are our models (CAE-FRs).}
\label{fig:neusiminsign}
\vspace{0.1in}
\end{figure*}

\begin{table*}[t]
\centering
\caption{The quantitative results of image reconstruction \ql{with constraints of} significant neurons and insignificant neurons in the region 3. The best result under different metrics is highlighted with boldface.}
\begin{tabular}{c|cc|cc|cc}
\hline
               Metric& \multicolumn{2}{c|}{MSE$\downarrow$} & \multicolumn{2}{c|}{SSIM$\uparrow$} & \multicolumn{2}{c}{PSNR$\uparrow$} \\ \hline
\diagbox{Model}{Significant?}  & $\checkmark$         & $\times$         & $\checkmark$         & $\times$          & $\checkmark$         & $\times$          \\ \hline
CAE-FR$_{1}$          & \textbf{0.043}       & 0.125      &\textbf{ 0.761  }     & 0.332       & \textbf{19.784   }   & 15.168      \\ 
CAE-FR$_{2}$         & \textbf{0.047}       & 0.082      & \textbf{0.743 }      & 0.547       & \textbf{19.467  }    & 16.970      \\ 
CAE-FR$_{3}$          & \textbf{0.049 }      & 0.116      & \textbf{0.724}       & 0.362       & \textbf{19.245 }     & 15.463      \\ 
CAE-FR$_{4}$ & 0.047       & \textbf{0.045  }    & 0.740       & \textbf{0.752 }      & 19.497      & \textbf{19.628}      \\ \hline
\end{tabular}
\label{tab:recon-results2}
\end{table*}

\begin{table*}[htbp]
\centering
\caption{Pearson correlation between the \ql{representations of} artificial and biological neurons in region 1, 2, and 3, respectively. The best result in each region under different layers of model is highlighted with boldface.}
\begin{adjustbox}{width=\columnwidth}
\begin{tabular}{c|cc|cc|cc|cc}
\hline
\diagbox{Region}{Model} & CNN-FR$_{1}$      & CAE-FR$_{1}$               & CNN-FR$_{2}$      &  CAE-FR$_{2}$               & CNN-FR$_{3}$      & CAE-FR$_{3}$  & CNN-FR$_{4}$      &  CAE-FR$_{4}$          \\ \hline
Region 1        & 0.341     & \textbf{0.346}    & 0.467     & \textbf{0.476}    & 0.454     & \textbf{0.455}  & 0.441     & \textbf{0.463}  \\ 
Region 2        & 0.257     & \textbf{0.281}    & 0.384     & \textbf{0.400}    & 0.333     & \textbf{0.338}  & 0.301     & \textbf{0.345}  \\ 
Region 3        & 0.246     & \textbf{0.260}    & 0.361     & \textbf{0.388}    & 0.372     & \textbf{0.404}  & 0.385     & \textbf{0.393}  \\ \hline
\end{tabular}
\end{adjustbox}
\label{tab:reuralsim}
\end{table*}
\subsection{Effects of neural responses in CAE-FR on image reconstruction}
\label{sec:results_imager_construction}


\ql{Ten examples of} reconstructed images by CAE-FR with neural responses in brain region 3 are presented in Fig.~\ref{fig:imgrecreg3}. \ql{It is obvious that CAE-FR models, no matter which layer the neural response is mapped to, can better reconstruct the original images compared with the standard CAE.} The results of the brain region 1 and 2 are illustrated in Appendix Fig.~\ref{fig:imgrecreg1} and Fig.~\ref{fig:imgrecreg2}. 
\ql{The quantitative comparisons of image reconstruction performance are listed} in the Table~\ref{tab:recon-results1}. It shows that CAE-FR$_{1}$, CAE-FR$_{4}$, and CAE-FR$_{2}$ achieve the best performance \ql{when the network gets information from the neurons in the regions 1, 2, and 3, respectively}. 




\ql{Fig.~\ref{fig:imgrecreg3} and Table~\ref{tab:recon-results1} suggests that the information from biological neurons could help CAE-FR for image reconstruction. Further questions are \textit{under what circumstances} and \textit{to what extent} CAE-FR is beneficial from the neural response. Our hypothesis is that \textit{only the biological neurons with high representation similarity to artificial neurons can provide information and contribute to CAE-FR}.}

To test our hypothesis, we identify the biological neurons that significantly correlate with the artificial neurons in the CAE-FR model, as well as the insignificant neurons in each brain region. Fig.~\ref{fig:imgrecinsginificant} shows the number of significant neurons and insignificant neurons in region 3 in the IR task. The numbers for region 1\&2 are in Appendix Fig.~\ref{fig:sigreg12}. Then, we train the CAE-FRs with the information from these two groups of neurons in region 3 for the IR task separately. The quantitative results of three metrics are shown in Table~\ref{tab:recon-results2}, confirming that the significant neurons largely improve the IR performance. In contrast, the insignificant neurons jeopardize CAE-FR, resulting in even worse IR performance (see results of Regions 3 in Table~\ref{tab:recon-results1}). These experiments verify our hypothesis, indicating that information from significant neurons can guide CAE-FR to better reconstruct images.

\subsection{Effects of image reconstruction in CAE-FR on neural representation similarity}
\label{sec:results_neural_similarity}

Here, we test the effects of image reconstruction in CAE-FR on neural representation similarity. The results of Pearson correlation coefficient between the neural representation of artificial neurons and of biological neurons in region 1, region 2 and region 3 are shown in Table~\ref{tab:reuralsim}. 
Our models (\ie CAE-FR$_i$) obtain larger PCC in all three regions compared to the baseline models without the image reconstruction loss (\ie CNN-FR$_i$). 
The results show that image reconstruction in our model favors neural representation similarity.
This phenomenon seems counter-intuitive as there is a tradeoff between the IR loss and the NSR loss in Eq.~\eqref{eq:loss-DAE-NR}. 
We hypothesize that \textit{the benefit may stem from feature learning for image reconstruction by CAE: more biological neurons would share representation similarity with artificial neurons when adding image reconstruction loss}. 
In fact, our experimental results verify the hypothesis (Fig.~\ref{fig:neusiminsign}), suggesting that image reconstruction loss can help CAE-FR to improve neural representation similarity between artificial neurons and biological neurons.

\subsection{Generalizability across variants of CNM and DAE}
\label{sec:results_imager_construction_resneu}
To investigate the generalizability of DAE-NR variants, we compare different DAE-NR variants with baseline CNMs and DAEs on IR and NRS tasks, respectively.

\textbf{Comparisons in IR task:} Table ~\ref{tab:recon-resultsdaenr} provides the quantitative comparison results of different models under various metrics of IR task in region 3. To test the effect of different CNMs for CAE, we compare the variants (\ie CAE-FR$_i$, CAE-FC$_i$, and CAE-FM$_i$) with the standard CAE on the IR task. The results show that the variants achieve the best performance when the network gets information from neurons in the region 3. Moreover, to examine generalizability in other DAE variants, we also implement some VAE variants (VAE-FR$_i$, VAE-FC$_i$,
and VAE-FM$_i$) and VQ-VAE variants (VQ-VAE-FR$_i$, VQ-VAE-FC$_i$, and VQ-VAE-FM$_i$) and compare their performance with the standard VAE and VQ-VAE on the IR task, respectively.
The results support our hypothesis that DAE-NR, VAE variants, and VQ-VAE variants have a better performance than baseline models without adding information from biological neurons.

\textbf{Comparisons in NSR task:} Pearson correlation coefficient (PCC$\uparrow$) between the neural representation of artificial neurons and biological neurons in region 1, region 2, and region 3 are shown in Table~\ref{tab:neusim-resultsdaenr}.  To verify the effect of different DAEs for CNN-FR, we compare the variants of CAE (CAE-FR$_i$, CAE-FC$_i$, and CAE-FM$_i$) with CNN-FR on the NSR task. Compared to the baseline models, the variants obtain larger PCC in all three regions with the constraints of image reconstruction loss. Furthermore, to validate the function of other CNM variants with different constraints of DAEs, we apply the variants of CNN-FC (CAE-FC$_i$, VAE-FC$_i$, and VQ-VAE-FC$_i$) and the variants of CNN-FM (CAE-FM$_i$, VAE-FM$_i$, and VQ-VAE-FM$_i$) to compare the performance on the NRS task, respectively. The results also support our hypothesis that the biologically-inspired variants of both CNN-FC and CNN-FM via our DAE-NR framework have a better performance than the original models.

Together, our experiments demonstrate that the DAE-NR has a good generalization in variants of CNM and DAE. We can add useful neural information to guide DAE in reconstructing visual input, and integrate image reconstruction loss into CNM to improve the representation similarity between artificial and biological neurons.

\begin{table*}[t]
\centering
\caption{Quantitative results of image reconstruction with all neurons in the region 3. The best results under different metrics for each region are highlighted in bold.}
\label{imgrecon}
\begin{adjustbox}{width=\columnwidth}
\begin{tabular}{c|ccc|ccc|ccc|ccc}
\hline
 $\textbf{h}_{i}$&
  \multicolumn{3}{c|}{$\textbf{h}_{1}$} &
  \multicolumn{3}{c|}{$\textbf{h}_{2}$} &
  \multicolumn{3}{c|}{$\textbf{h}_{3}$} &
  \multicolumn{3}{c}{$\textbf{h}_{4}$} \\ \hline
\diagbox{Model}{Metric} &
  MSE$\downarrow$ &
  PSNR$\uparrow$&
  SSIM$\uparrow$&
  MSE$\downarrow$ &
  PSNR$\uparrow$ &
  SSIM$\uparrow$&
  MSE$\downarrow$ &
  PSNR$\uparrow$ &
  SSIM$\uparrow$&
  MSE$\downarrow$ &
  PSNR$\uparrow$ &
  SSIM$\uparrow$ \\ \hline
\multicolumn{1}{c|}{CAE$_{i}$} &
  \multicolumn{1}{c}{0.074} &
  \multicolumn{1}{c}{17.289} &
  \multicolumn{1}{c|}{0.593} &
  \multicolumn{1}{c}{0.074} &
  \multicolumn{1}{c}{17.289} &
  \multicolumn{1}{c|}{0.593} &
  \multicolumn{1}{c}{0.074} &
  \multicolumn{1}{c}{17.289} &
  \multicolumn{1}{c|}{0.593}&
  \multicolumn{1}{c}{0.074} &
  \multicolumn{1}{c}{17.289} &
  \multicolumn{1}{c}{0.593} \\ 
CAE-FR$_{i}$ \textbf{(Ours)} & 
  \textbf{0.043} &
  19.832 &
  0.764 &
  \textbf{0.043} &
  \textbf{19.852} &
  \textbf{0.767} &
  0.045 &
  19.698 &
  0.758 &
  0.051 &
  19.107 &
  0.718 \\ 
CAE-FC$_{i}$~\textbf{(Ours)}  &
  \textbf{0.043} &
  \textbf{19.876} &
  \textbf{0.768} &
  0.044 &
  19.811 &
  0.763 &
  0.061 &
  18.290 &
  0.734 &
  \textbf{0.044} &
  \textbf{19.802} &
  \textbf{0.760} \\ 
CAE-FM$_{i}$~\textbf{(Ours)} &
  0.045 &
  19.692 &
 0.761 &
  \textbf{0.043} &
  19.843 &
  0.764 &
  \textbf{0.044} &
  \textbf{19.818} &
  \textbf{0.761} &
  \textbf{0.044} &
  19.757 &
  0.757 \\  \hline
VAE$_{i}$ &
  0.104 &
  15.983 &
  0.402 &
  0.104 &
  15.983 &
  0.402 &
  0.104 &
  15.983 &
  0.402 &
  0.104 &
  15.983 &
  0.402\\ 
VAE-FR$_{i}$~\textbf{(Ours)} &
  \textbf{0.101} &
  \textbf{16.106} &
  0.414 &
  \textbf{0.101} &
  16.066 &
  0.409 &
  \textbf{0.101} &
  16.083 &
  \textbf{0.417} &
  0.102 &
  16.063 &
  0.414 \\ 
VAE-FC$_{i}$~\textbf{(Ours)} &
  0.103 &
  16.003 &
  0.412 &
  \textbf{0.101} &
  \textbf{16.094} &
  0.412 &
  \textbf{0.101} &
  \textbf{16.096} &
  0.416 &
  \textbf{0.101} &
  \textbf{16.089} &
  \textbf{0.418} \\ 
VAE-FM$_{i}$~\textbf{(Ours)} &
  0.102 &
  16.046 &
  \textbf{0.415} &
  \textbf{0.101} &
  16.073 &
  \textbf{0.414} &
  0.102 &
  16.025 &
  0.413 &
  0.103 &
  16.017 &
  0.412 \\ \hline
VQ-VAE$_{i}$ &
  0.082 &
  16.959 &
  0.497 &
  0.082 &
  16.959 &
  0.497 &
  0.082 &
  16.959 &
  0.497 &
  0.082 &
  16.959 &
  0.497 \\ 
VQ-VAE-FR$_{i}$~\textbf{(Ours)} &
  \textbf{0.071} &
  17.636 &
  0.569 &
  0.069 &
  17.741 &
  0.568 &
  0.073 &
  17.487 &
  \textbf{0.557} &
  0.071 &
  17.581 &
  0.560 \\ 
VQ-VAE-FC$_{i}$~\textbf{(Ours)} &
  0.076 &
  17.264 &
  0.526 &
  0.075 &
  17.324 &
  0.525 &
  \textbf{0.072} &
  \textbf{17.539} &
  0.547 &
  \textbf{0.067} &
  \textbf{17.839} &
  \textbf{0.581} \\ 
VQ-VAE-FM$_{i}$~\textbf{(Ours)} &
  0.068 &
  \textbf{17.782} &
  \textbf{0.572} &
  \textbf{0.068} &
  \textbf{17.791} &
  \textbf{0.571} &
  0.091 &
  16.496 &
  0.546 &
  0.072 &
  17.508 &
  0.542 \\ \hline
\end{tabular}
\end{adjustbox}
\label{tab:recon-resultsdaenr}
\end{table*}

\begin{table*}[h]
\centering
\caption{The results of PCC$\uparrow$ between the representations of artificial and biological neurons in three brain regions. The best results under different metrics for each region are highlighted in bold.}
\label{neusim}
\begin{adjustbox}{width=\columnwidth}
\begin{tabular}{c|cccc|cccc|cccc}
\hline
 Region &
  \multicolumn{4}{c|}{Region 1} &
  \multicolumn{4}{c|}{Region 2} &
  \multicolumn{4}{c}{Region 3} \\ \hline
\diagbox{Model}{$\textbf{h}_{i}$} &
  $\textbf{h}_{1}$&
  $\textbf{h}_{2}$&
  $\textbf{h}_{3}$&
  $\textbf{h}_{4}$&
  $\textbf{h}_{1}$&
  $\textbf{h}_{2}$&
  $\textbf{h}_{3}$&
  $\textbf{h}_{4}$&
  $\textbf{h}_{1}$&
  $\textbf{h}_{2}$&
  $\textbf{h}_{3}$&
  $\textbf{h}_{4}$\\ \hline
\multicolumn{1}{c|}{CNN-FR$_{i}$} &
  \multicolumn{1}{c}{0.335} &
  \multicolumn{1}{c}{0.455} &
  \multicolumn{1}{c}{0.432} &
  \multicolumn{1}{c|}{0.431} &
  \multicolumn{1}{c}{0.254} &
  \multicolumn{1}{c}{0.375} &
  \multicolumn{1}{c}{0.309} &
  \multicolumn{1}{c|}{0.305} &
  \multicolumn{1}{c}{0.244} &
  \multicolumn{1}{c}{0.348} &
  \multicolumn{1}{c}{0.350} &
  \multicolumn{1}{c}{0.359} \\ 
CAE-FR$_{i}$~\textbf{(Ours)} & 
  0.338 &
  0.471 &
  0.449 &
  \textbf{0.452} &
  0.277 &
  \textbf{0.396} &
  0.319 &
  0.348 &
  0.254 &
  0.377 &
  \textbf{0.397} &
  0.394 \\ 
VAE-FR$_{i}$~\textbf{(Ours)} &
  0.338 &
  \textbf{0.479} &
  0.455 &
  0.450 &
  0.277 &
  0.386 &
  0.317 &
  0.342 &
  \textbf{0.269} &
  0.375 &
  0.382 &
  \textbf{0.404} \\ 
VQ-VAE-FR$_{i}$~\textbf{(Ours)} &
  \textbf{0.349} &
  0.470 &
  \textbf{0.463} &
  0.444 &
  \textbf{0.295} &
  0.388 &
  \textbf{0.337} &
  \textbf{0.349} &
  0.261 &
  \textbf{0.394} &
  0.382 &
  0.387 \\  \hline
CNN-FC$_{i}$ &
  0.343 &
  0.349 &
  0.397 &
  0.435 &
  0.247 &
  0.268 &
  0.292 &
  0.328 &
  0.325 &
  0.316 &
  0.344 &
  0.374 \\ 
CAE-FC$_{i}$~\textbf{(Ours)} &
  \textbf{0.349} &
  0.382 &
  0.415 &
  \textbf{0.438} &
  \textbf{0.266} &
  0.281 &
  0.319 &
  0.341 &
  0.335 &
  \textbf{0.341} &
  0.384 &
  0.399 \\ 
VAE-FC$_{i}$~\textbf{(Ours)} &
  0.340 &
  \textbf{0.383} &
  \textbf{0.418} &
  \textbf{0.438} &
  0.255 &
  \textbf{0.290} &
  \textbf{0.321} &
  0.343 &
  0.337 &
  0.337 &
  0.379 &
  0.398 \\ 
VQ-VAE-FC$_{i}$~\textbf{(Ours)} &
  0.347 &
  0.377 &
  0.415 &
  0.434 &
  0.264 &
  0.289 &
  0.313 &
  \textbf{0.361} &
 \textbf{0.338} &
  \textbf{0.341} &
  \textbf{0.388} &
  \textbf{0.427} \\ \hline
CNN-FM$_{i}$ &
  0.150 &
  0.222 &
  0.110 &
  0.220 &
  0.123 &
  0.186 &
  0.127 &
  0.162 &
  0.064 &
  0.140 &
  0.102 &
  0.178 \\ 
CAE-FM$_{i}$~\textbf{(Ours)} &
  \textbf{0.154} &
  0.235 &
  0.115 &
  0.227 &
  0.134 &
  0.222 &
  \textbf{0.144} &
  \textbf{0.182} &
  0.073 &
  0.171 &
  0.109 &
 \textbf{0.203} \\ 
VAE-FM$_{i}$~\textbf{(Ours)} &
  0.149 &
  0.238 &
  \textbf{0.125} &
  \textbf{0.240} &
  \textbf{0.136} &
  \textbf{0.224} &
  0.131 &
  0.176 &
  \textbf{0.074} &
  \textbf{0.172} &
  \textbf{0.111} &
  0.201 \\ 
VQ-VAE-FM$_{i}$~\textbf{(Ours)} &
  0.153 &
  \textbf{0.239} &
  0.115 &
  0.221 &
  0.135 &
  0.221 &
  0.136 &
  0.162 &
  0.069 &
  0.162 &
  \textbf{0.111} &
  0.200 \\ \hline
\end{tabular}
\end{adjustbox}
\label{tab:neusim-resultsdaenr}
\end{table*}

\section{Conclusion}
In this study, we proposed the DAE-NRs, a hybrid framework that integrates the neural response into deep auto-encoder models. Inspired by CNN-based neural activity prediction models in computational neuroscience~\cite{klindt2017neural,bashivan2019neural}, we used a Poisson loss to bring neural information to a specific layer of DAE, resulting in better image reconstruction performance. In return, the IR task facilitates feature learning in DAE and leads to higher neural representation similarity between biological neurons and artificial neurons. Our work bridges the gap between DAE for image reconstruction and CNM for neural representation similarity.


\paragraph{Broader impact}  Besides the IR task and NRS task, DAE-NR could enable many other potential applications in the future. For instance, DAE-NR provides a more natural way to synthesize images that maximize or control neural activity, compared with the method proposed by~\cite{bashivan2019neural}. Also, DAE-NR can serve as a data engine for biological experiments, synthesizing neural responses to visual stimuli.
Moreover, it is important to investigate the generalizability of DAE-NR on variants of auto-encoder~\cite{larsen2016autoencoding,khattar2019mvae,park2020swapping,wang2021encoder,cai2021unified}, on other types of stimuli (\eg, sound and face) and in other tasks (\eg classification, generation and detection~\cite{ran2021detecting}).
Although we only tested DAE-NR with mice neural data, the DAE-NR framework can be easily extended to the neurons in primates (\eg monkey~\cite{Zhang2021VisualAI} or humans). 
DAE-NR opens a new window for combining artificial intelligence and brain intelligence.




\newpage
\appendix
\section{Experimental settings}
\begin{figure*}[hbp]
\centering
\includegraphics[width=0.8\textwidth]{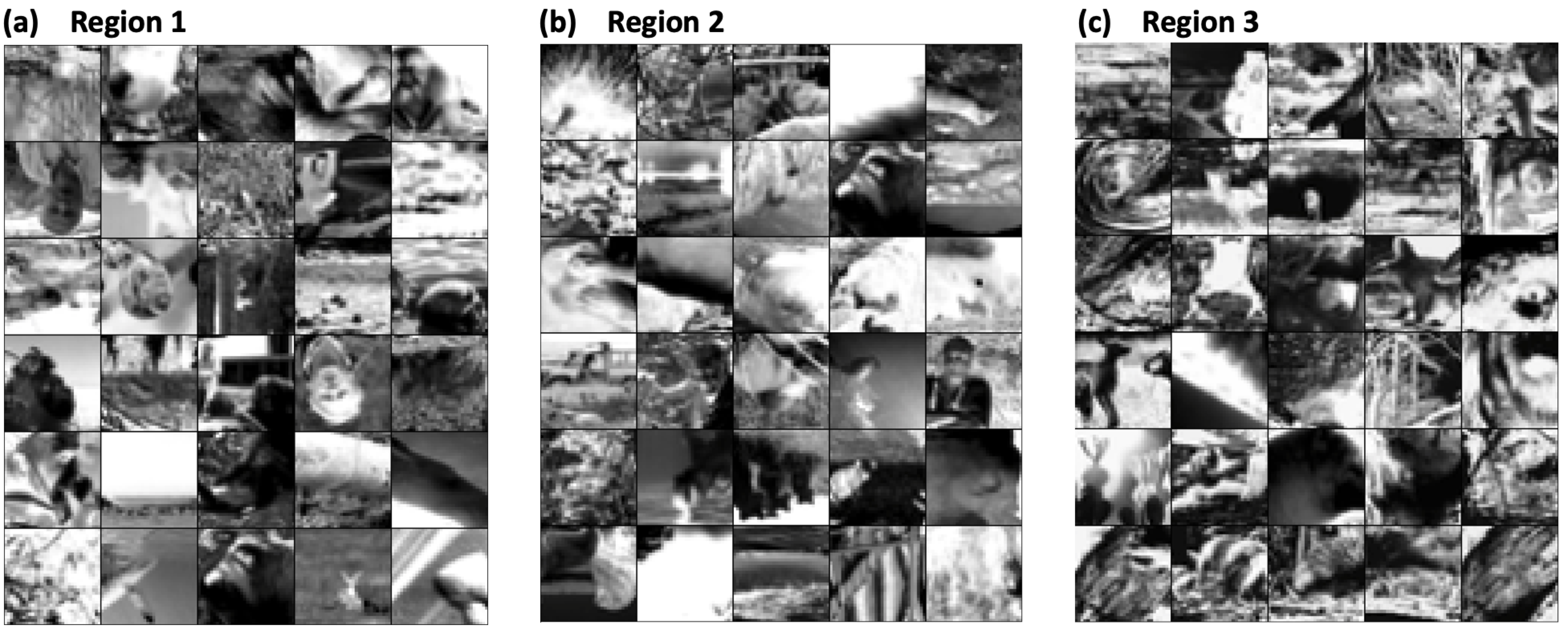}
\caption{The \ql{examples of} training stimuli \ql{in} each brain region. }
\label{fig:trnimg}
\vspace{0.1in}
\end{figure*}

In this section, we introduce the details of the settings in our experiments. Firstly, Table~\ref{tab:numregs} is the number of neural datasets from three brain regions and Fig.~\ref{fig:trnimg} illustrates the training image of the neural stimuli. Secondly, the Table~\ref{tab:setexpimgrec} and Table~\ref{tab:setexpneusim} are the settings of the hyperparameters for image reconstruction and neural similarity experiments, respectively.
Thirdly, the Table ~\ref{tab:setexpimgrecdarvari3} is the setting of the hyperparameters of  different DAE-NR variants for image reconstruction
experiments for image reconstruction on region 3 . Fourthly, the Table~\ref{tab:setexpneusimreg1},  Table~\ref{tab:setexpneusimreg2} and Table~\ref{tab:setexpneusimreg3} are the settings the hyperparameters of different DAE-NR variants for neural similarity
experiments on Region 1,2 and 3, respectively. 

\begin{table*}[hbp]
\centering  
\caption{The neural dataset \ql{containing neural responses in} three brain regions \ql{under visual stimulation}~\cite{Antolk2016ModelCB}. The number in bracket is the repeated times.}  
\begin{tabular}{c|c|c|c}  
	\hline  
& Number of train images & Number of test images & Number of neurons \\  
\hline
Region 1 & 1800 (1) & 50 (10) & 103	\\

Region 2 &1260 (1) &50 (8) & 55 \\

Region 3 &1800 (1) &50 (12) & 102 \\
\hline
\end{tabular}
\label{tab:numregs}  %
\end{table*}

\begin{table*}[hbp]
\centering
\caption{The settings of the hyperparameters for image reconstruction experiments.}
\begin{tabular}{c|c|c|c}
\hline
                               & Region 1       & Region 2       & Region 3       \\ \hline
Models & $\alpha:\beta$ & $\alpha:\beta$ & $\alpha:\beta$ \\ \hline
CAE-FR$_{1}$                          & 1e-0:1e-5      & 1e-0:1e-3      & 1e-0:1e-4      \\ 
CAE-FR$_{2}$                         & 1e-0:1e-5      & 1e-0:1e-4      & 1e-0:1e-3      \\ 
CAE-FR$_{3}$                          & 1e-0:1e-4      & 1e-0:1e-3      & 1e-0:1e-5      \\ 
CAE-FR$_{4}$                          & 1e-0:1e-5      & 1e-0:1e-4      & 1e-0:1e-4      \\ \hline
\end{tabular}`
\label{tab:setexpimgrec}
\end{table*}

\begin{table*}[hbp]
\centering
\caption{The settings the hyperparameters for neural similarity experiments.}

\begin{tabular}{c|c|c|c}
\hline
      & Region 1  & Region 2  & Region 3  \\ \hline
Models & $\alpha:\beta$ & $\alpha:\beta$ & $\alpha:\beta$ \\ \hline
CAE-FR$_{1}$ & 1e-0:9e-1 & 7e-1:1e-0 & 8e-1:1e-0 \\ 
CAE-FR$_{2}$ & 7e-1:1e-0 & 1e-0:1e-2 & 4e-1:1e-0 \\ 
CAE-FR$_{3}$ & 1e-0:6e-1 & 1e-4:1e-0 & 1e-0:4e-1 \\ 
CAE-FR$_{4}$ & 1e-0:9e-1 & 1e-0:1e-2 & 1e-0:9e-1 \\ \hline
\end{tabular}
\label{tab:setexpneusim}
\end{table*}

\begin{table*}[hbp]
\centering
\caption{The settings the hyperparameters of different DAE-NR variants for image reconstruction experiments on region 3.  }
\label{tab:setexpimgrecdarvari3}
\begin{tabular}{c|c|c|c|c}
\hline
      & $h_1$  & $h_2$  & $h_3$  & $h_4$ \\ \hline
Models & $\alpha:\beta$ & $\alpha:\beta$ & $\alpha:\beta$ & $\alpha:\beta$ \\ \hline
CAE-FR$_{}$ & 1e-0:1e-4 & 1e-0:5e-3 & 1e-0:1e-3 & 1e-0:1e-3\\ 
CAE-FC$_{}$ & 1e-0:1e-4 & 1e-0:1e-4 & 1e-0:1e-4 & 1e-0:1e-4 \\ 
CAE-FM$_{}$ & 1e-0:2e-1 & 1e-0:1e-3 & 1e-0:1e-4 & 1e-0:1e-4\\ \hline
VAE-FR$_{}$ & 1e-0:2e-2 & 1e-0:1e-2 & 1e-0:2e-1 & 1e-0:5e-3 \\
VAE-FC$_{}$ & 1e-0:1e-2 & 1e-0:5e-3 & 1e-0:1e-3 & 1e-0:1e-1 \\ 
VAE-FM$_{}$ & 1e-0:5e-3 & 1e-0:5e-2 & 1e-0:2e-1 & 1e-0:5e-3\\ \hline
VQ-VAE-FR$_{}$ & 1e-0:2e-2 & 1e-0:2e-1 & 1e-0:1e-2 & 1e-0:1e-3\\ 
VQ-VAE-FC$_{}$ & 1e-0:1e-3 & 1e-0:1e-4 & 1e-0:1e-4 & 1e-0:1e-4\\
VQ-VAE-FM$_{}$ & 1e-0:2e-1 & 1e-0:1e-3 & 1e-0:1e-2 & 1e-0:1e-3\\ \hline
\end{tabular}
\end{table*}

\begin{table*}[hbp]
\centering
\caption{The settings the hyperparameters of different DAE-NR variants for neural similarity experiments on Region 1  }

\begin{tabular}{c|c|c|c|c}
\hline
      & $h_1$  & $h_2$  & $h_3$  & $h_4$ \\ \hline
Models & $\alpha:\beta$ & $\alpha:\beta$ & $\alpha:\beta$ & $\alpha:\beta$\\ \hline
CAE-FR$_{}$ & 1e-0:5e-1 & 1e-0:5e-2 & 1e-0:5e-2 & 1e-0:1e-2 \\ 
VAE-FR$_{}$ & 1e-3:1e-0 & 2e-2:1e-0 & 1e-3:1e-0 & 1e-1:1e-0\\ 
VQ-VAE-FR$_{}$ & 1e-0:2e-2 & 1e-0:2e-1 & 1e-0:5e-3 & 1e-0:2e-1 \\ \hline
CAE-FC$_{}$ & 5e-3:1e-0 & 1e-3:1e-0 & 2e-2:1e-0 & 1e-2:1e-0 \\
VAE-FC$_{}$ & 2e-2:1e-0 & 1e-0:1e-2 & 2e-1:1e-0 & 1e-3:1e-0 \\
VQ-VAE-FC$_{}$ & 1e-0:1e-1 & 1e-0:5e-1 & 1e-4:1e-0 & 1e-2:1e-0 \\\hline
CAE-FM$_{}$ & 1e-2:1e-0 & 5e-1:1e-0 & 1e-0:1e-1 & 1e-0:5e-1 \\
VAE-FM$_{}$ & 1e-4:1e-0 & 1e-3:1e-0 & 1e-1:1e-0 & 1e-1:1e-0 \\
VQ-VAE-FM$_{}$ & 1e-1:1e-0 & 1e-1:1e-0 & 1e-0:1e-2 & 1e-0:1e-0 \\
\hline
\end{tabular}
\label{tab:setexpneusimreg1}
\end{table*}

\begin{table*}[hbp]
\centering
\caption{The settings the hyperparameters of different DAE-NR variants for neural similarity experiments on Region 2 }

\begin{tabular}{c|c|c|c|c}
\hline
      & $h_1$  & $h_2$  & $h_3$  & $h_4$ \\ \hline
Models & $\alpha:\beta$ & $\alpha:\beta$ & $\alpha:\beta$ & $\alpha:\beta$\\ \hline
CAE-FR$_{}$ & 1e-0:2e-1 & 1e-0:2e-2 & 5e-3:1e-0 & 1e-0:5e-3\\ 
VAE-FR$_{}$ & 5e-3:1e-0 & 2e-1:1e-0 & 5e-2:1e-0 & 5e-1:1e-0\\ 
VQ-VAE-FR$_{}$ & 1e-0:1e-1 & 2e-1:1e-0 & 1e-0:1e-2& 1e-0:2e-1 \\ \hline
CAE-FC$_{}$ & 1e-2:1e-0 & 5e-3:1e-0 & 1e-0:5e-2 & 1e-4:1e-0\\
VAE-FC$_{}$ & 1e-2:1e-0 & 5e-3:1e-0 & 1e-0:5e-1 & 1e-4:1e-0\\
VQ-VAE-FC$_{}$ & 1e-0:1e-0 & 5e-3:1e-0 & 1e-3:1e-0& 1e-3:1e-0 \\\hline
CAE-FM$_{}$ & 1e-0:1e-1 & 1e-0:1e-2 & 1e-0:5e-3 & 1e-0:5e-3\\
VAE-FM$_{}$ & 5e-2:1e-0 & 1e-4:1e-0 & 5e-3:1e-0 & 2e-2:1e-0\\
VQ-VAE-FM$_{}$ & 1e-0:5e-2 & 5e-2:1e-0 & 1e-0:1e-1 & 1e-0:5e-1\\
\hline
\end{tabular}
\label{tab:setexpneusimreg2}
\end{table*}

\begin{table*}[hbp]
\centering
\caption{The settings the hyperparameters of different DAE-NR variants for neural similarity experiments on Region 3  }
\label{tab:setexpneusimreg3}
\begin{tabular}{c|c|c|c|c}
\hline
      & $h_1$  & $h_2$  & $h_3$  & $h_4$ \\ \hline
Models & $\alpha:\beta$ & $\alpha:\beta$ & $\alpha:\beta$ & $\alpha:\beta$\\ \hline
CAE-FR$_{}$ & 1e-0:1e-1 & 1e-0:1e-0 & 1e-0:5e-2 & 1e-0:1e-2 \\
VAE-FR$_{}$ & 1e-0:1e-4 & 5e-1:1e-0 & 1e-2:1e-0 & 1e-0:1e-0 \\
VQ-VAE-FR$_{}$ & 1e-0:2e-1 & 1e-0:2e-2 & 1e-0:1e-0 & 1e-4:1e-0 \\ \hline
CAE-FC$_{}$ & 1e-0:2e-1 & 5e-2:1e-0 & 1e-0:1e-1 & 1e-0:2e-1 \\
VAE-FC$_{}$ & 1e-3:1e-0 & 1e-2:1e-0 & 5e-3:1e-0& 2e-1:1e-0  \\
VQ-VAE-FC$_{}$ & 1e-1:1e-0 & 1e-3:1e-0 & 1e-2:1e-0& 1e-4:1e-0  \\  \hline
CAE-FM$_{}$ & 5e-3:1e-0 & 1e-0:5e-3 & 1e-0:5e-1 & 1e-2:1e-0 \\
VAE-FM$_{}$ & 1e-3:1e-0 & 1e-0:2e-1 & 2e-2:1e-0 & 5e-2:1e-0 \\
VQ-VAE-FM$_{4}$ & 1e-0:5e-1 & 1e-4:1e-0 & 1e-3:1e-0& 1e-3:1e-0  \\
\hline
\end{tabular}
\end{table*}

\section{Additional experiments}
\begin{figure*}[hbp]
    \centering
    \includegraphics[width=0.8\textwidth]{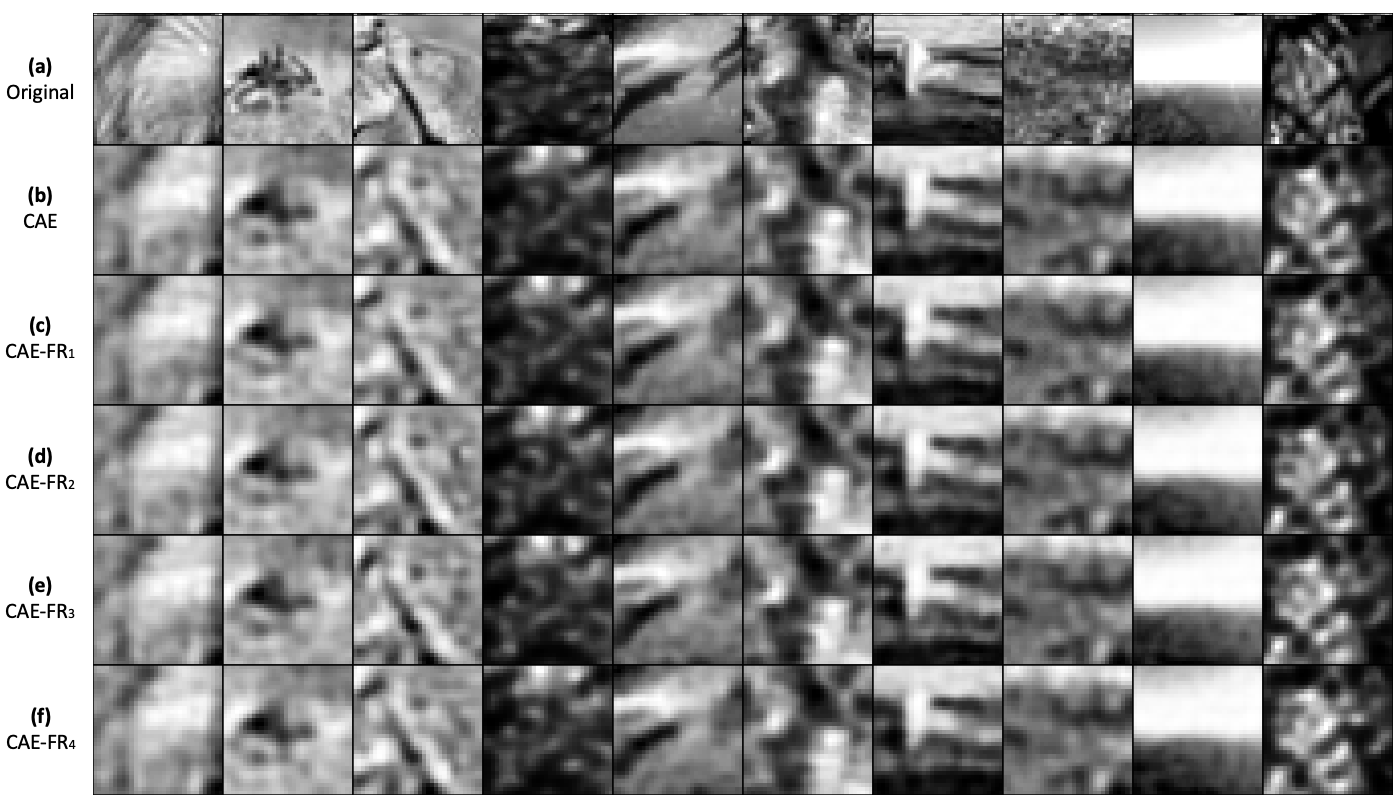}
    \caption{The \ql{reconstructed images with neurons in Region 1}. \ql{From top to bottom,} each row displays the original images (a), the images reconstructed by CAE (b), CAE-FR$_{1}$ (c), CAE-F$_{2}$ (d), CAE-F$_{3}$ (e), CAE-FR$_{4}$ (f), respectively.}
    \label{fig:imgrecreg1}
\end{figure*}

\begin{figure*}[hbp]
    \centering
    \includegraphics[width=0.8\textwidth]{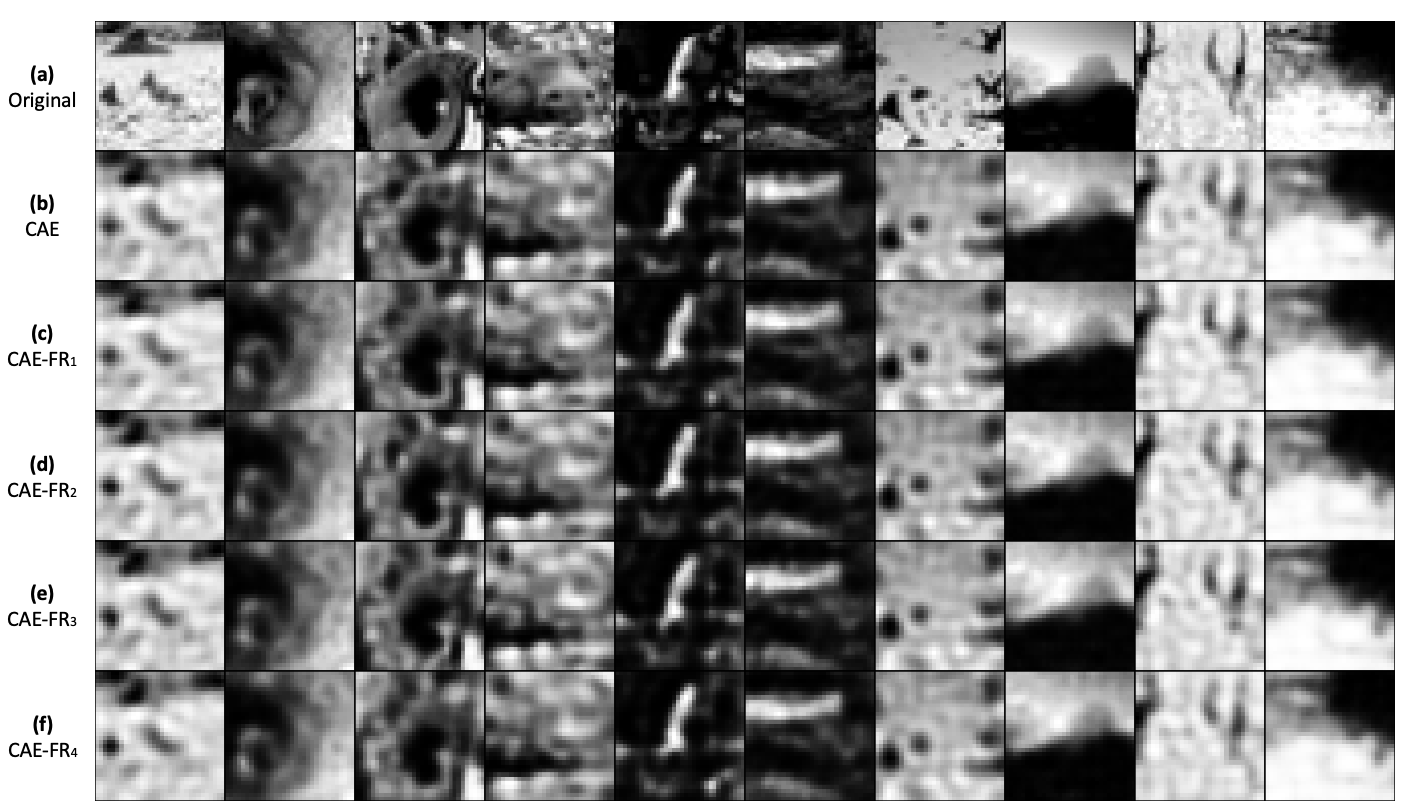}
    \caption{The \ql{reconstructed images with neurons in Region 2}. \ql{From top to bottom,} each row displays the original images (a), the images reconstructed by CAE (b), CAE-FR$_{1}$ (c), CAE-F$_{2}$ (d), CAE-F$_{3}$ (e), CAE-FR$_{4}$ (f), respectively.}
    \label{fig:imgrecreg2}
\end{figure*}

\begin{figure*}[t]
\centering
\subfloat{
    \includegraphics[page=1,width=2.5in]{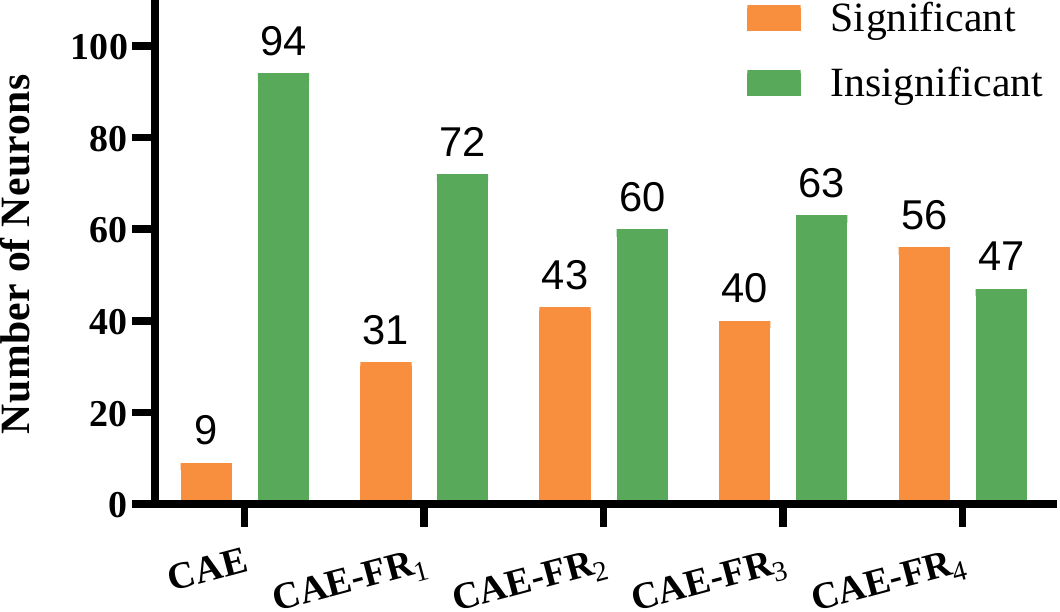}}
\subfloat{
    \includegraphics[page=1,width=2.5in]{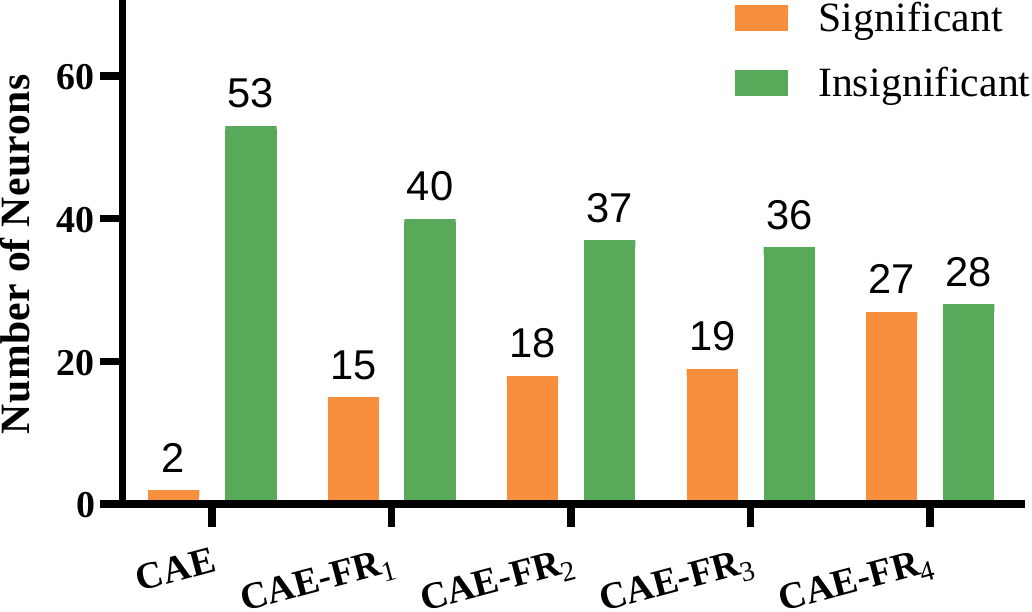}}
\caption{The number of significant \ql{neurons} and insignificant neurons in region 1 and 2 in the image reconstruction experiments. \ql{The threshold for significance is $p \leq 0.05$.}}
\label{fig:sigreg12}
\vspace{0.1in}
\end{figure*}

In this section, we provide results from additional experiments. Fig.~\ref{fig:imgrecreg1} and Fig.~\ref{fig:imgrecreg2} display the results of image reconstruction using the neural responses on the region 1 and 2. Fig.~\ref{fig:sigreg12} shows the number of significant neurons and insignificant neurons of regions 1 and 2 in the image reconstruction experiments, respectively.

{\small
\begin{thebibliography}{10}

\bibitem{Antolk2016ModelCB}
J.~Antol{\'i}k, S.~Hofer, J.~Bednar, and T.~Mrsic-Flogel.
\newblock Model constrained by visual hierarchy improves prediction of neural
  responses to natural scenes.
\newblock {\em PLoS Computational Biology}, 12, 2016.

\bibitem{bashivan2019neural}
P.~Bashivan, K.~Kar, and J.~DiCarlo.
\newblock Neural population control via deep image synthesis.
\newblock {\em Science}, 364, 2019.

\bibitem{Cadena2019DeepCM}
S.~A. Cadena, G.~H. Denfield, E.~Y. Walker, L.~A. Gatys, A.~Tolias, M.~Bethge,
  and A.~S. Ecker.
\newblock Deep convolutional models improve predictions of macaque v1 responses
  to natural images.
\newblock {\em PLoS Computational Biology}, 15, 2019.

\bibitem{cai2021unified}
Y.~Cai, Y.~Wang, Y.~Zhu, T.-J. Cham, J.~Cai, J.~Yuan, J.~Liu, C.~Zheng, S.~Yan,
  H.~Ding, et~al.
\newblock A unified 3d human motion synthesis model via conditional variational
  auto-encoder.
\newblock In {\em Proceedings of the IEEE/CVF International Conference on
  Computer Vision}, pages 11645--11655, 2021.

\bibitem{Carandini2005DoWK}
M.~Carandini, J.~B. Demb, V.~Mante, D.~J. Tolhurst, Y.~Dan, B.~A. Olshausen,
  J.~L. Gallant, and N.~C. Rust.
\newblock Do we know what the early visual system does?
\newblock {\em The Journal of Neuroscience}, 25:10577 -- 10597, 2005.

\bibitem{das2019computational}
G.~P. Das, P.~J. Vance, D.~Kerr, S.~A. Coleman, T.~M. McGinnity, and J.~K. Liu.
\newblock Computational modelling of salamander retinal ganglion cells using
  machine learning approaches.
\newblock {\em Neurocomputing}, 325:101--112, 2019.

\bibitem{DiCarlo2012HowDT}
J.~DiCarlo, D.~Zoccolan, and N.~Rust.
\newblock How does the brain solve visual object recognition?
\newblock {\em Neuron}, 73:415--434, 2012.

\bibitem{Drger1975ReceptiveFO}
U.~C. Dr{\"a}ger.
\newblock Receptive fields of single cells and topography in mouse visual
  cortex.
\newblock {\em Journal of Comparative Neurology}, 160, 1975.

\bibitem{Federer2020ImprovedOR}
C.~Federer, H.~Xu, A.~Fyshe, and J.~Zylberberg.
\newblock Improved object recognition using neural networks trained to mimic
  the brain's statistical properties.
\newblock {\em Neural networks : the official journal of the International
  Neural Network Society}, 131:103--114, 2020.

\bibitem{Froudarakis2014PopulationCI}
E.~Froudarakis, P.~Berens, A.~S. Ecker, R.~J. Cotton, F.~H. Sinz, D.~Yatsenko,
  P.~Saggau, M.~Bethge, and A.~S. Tolias.
\newblock Population code in mouse v1 facilitates read-out of natural scenes
  through increased sparseness.
\newblock {\em Nature neuroscience}, 17:851 -- 857, 2014.

\bibitem{Fukushima1983NeocognitronAN}
K.~Fukushima, S.~Miyake, and T.~Ito.
\newblock Neocognitron: A neural network model for a mechanism of visual
  pattern recognition.
\newblock {\em IEEE Transactions on Systems, Man, and Cybernetics},
  SMC-13:826--834, 1983.

\bibitem{goodfellow2016deep}
I.~Goodfellow, Y.~Bengio, and A.~Courville.
\newblock {\em Deep learning}.
\newblock MIT press, 2016.

\bibitem{hinton2006reducing}
G.~E. Hinton and R.~R. Salakhutdinov.
\newblock Reducing the dimensionality of data with neural networks.
\newblock {\em science}, 313(5786):504--507, 2006.

\bibitem{hubel1962receptive}
D.~H. Hubel and T.~N. Wiesel.
\newblock Receptive fields, binocular interaction and functional architecture
  in the cat's visual cortex.
\newblock {\em The Journal of physiology}, 160(1):106--154, 1962.

\bibitem{Hubel1968ReceptiveFA}
D.~H. Hubel and T.~N. Wiesel.
\newblock Receptive fields and functional architecture of monkey striate
  cortex.
\newblock {\em The Journal of Physiology}, 195, 1968.

\bibitem{khattar2019mvae}
D.~Khattar, J.~S. Goud, M.~Gupta, and V.~Varma.
\newblock Mvae: Multimodal variational autoencoder for fake news detection.
\newblock In {\em The world wide web conference}, pages 2915--2921, 2019.

\bibitem{Kingma2014AutoEncodingVB}
D.~P. Kingma and M.~Welling.
\newblock Auto-encoding variational bayes.
\newblock {\em International Conference on Learning Representations(ICLR)},
  2014.

\bibitem{klindt2017neural}
D.~A. Klindt, A.~S. Ecker, T.~Euler, and M.~Bethge.
\newblock Neural system identification for large populations separating what
  and where.
\newblock In {\em Advances in Neural Information Processing Systems (NeurIPS)},
  pages 3509--3519, 2017.

\bibitem{Krizhevsky2012ImageNetCW}
A.~Krizhevsky, I.~Sutskever, and G.~E. Hinton.
\newblock Imagenet classification with deep convolutional neural networks.
\newblock {\em Communications of the ACM}, 60:84 -- 90, 2012.

\bibitem{larsen2016autoencoding}
A.~B.~L. Larsen, S.~K. S{\o}nderby, H.~Larochelle, and O.~Winther.
\newblock Autoencoding beyond pixels using a learned similarity metric.
\newblock In {\em International conference on machine learning}, pages
  1558--1566. PMLR, 2016.

\bibitem{LeCun1989HandwrittenDR}
Y.~LeCun, B.~Boser, J.~Denker, D.~Henderson, R.~Howard, W.~Hubbard, and
  L.~Jackel.
\newblock Handwritten digit recognition with a back-propagation network.
\newblock In {\em Advances in Neural Information Processing Systems (NeurIPS)},
  1989.

\bibitem{macpherson2021natural}
T.~Macpherson, A.~Churchland, T.~Sejnowski, J.~DiCarlo, Y.~Kamitani,
  H.~Takahashi, and T.~Hikida.
\newblock Natural and artificial intelligence: A brief introduction to the
  interplay between ai and neuroscience research.
\newblock {\em Neural Networks}, 2021.

\bibitem{masci2011stacked}
J.~Masci, U.~Meier, D.~Cire{\c{s}}an, and J.~Schmidhuber.
\newblock Stacked convolutional auto-encoders for hierarchical feature
  extraction.
\newblock In {\em International conference on artificial neural networks},
  pages 52--59. Springer, 2011.

\bibitem{mcintosh2016deep}
L.~McIntosh, N.~Maheswaranathan, A.~Nayebi, S.~Ganguli, and S.~Baccus.
\newblock Deep learning models of the retinal response to natural scenes.
\newblock {\em Advances in neural information processing systems(NeurIPS)},
  29:1369--1377, 2016.

\bibitem{meyer2017models}
A.~F. Meyer, R.~S. Williamson, J.~F. Linden, and M.~Sahani.
\newblock Models of neuronal stimulus-response functions: elaboration,
  estimation, and evaluation.
\newblock {\em Frontiers in systems neuroscience}, 10:109, 2017.

\bibitem{Niell2008HighlySR}
C.~M. Niell, M.~P. Stryker, and W.~M. Keck.
\newblock Highly selective receptive fields in mouse visual cortex.
\newblock {\em The Journal of Neuroscience}, 28:7520 -- 7536, 2008.

\bibitem{park2020swapping}
T.~Park, J.-Y. Zhu, O.~Wang, J.~Lu, E.~Shechtman, A.~Efros, and R.~Zhang.
\newblock Swapping autoencoder for deep image manipulation.
\newblock {\em Advances in Neural Information Processing Systems},
  33:7198--7211, 2020.

\bibitem{ran2021detecting}
X.~Ran, M.~Xu, L.~Mei, Q.~Xu, and Q.~Liu.
\newblock Detecting out-of-distribution samples via variational auto-encoder
  with reliable uncertainty estimation.
\newblock {\em Neural Networks}, 2021.

\bibitem{Ravishankar2020ImageRF}
S.~Ravishankar, J.~C. Ye, and J.~A. Fessler.
\newblock Image reconstruction: From sparsity to data-adaptive methods and
  machine learning.
\newblock {\em Proceedings of the IEEE}, 108:86--109, 2020.

\bibitem{Rowekamp2017CrossorientationSI}
R.~J. Rowekamp and T.~Sharpee.
\newblock Cross-orientation suppression in visual area v2.
\newblock {\em Nature Communications}, 8, 2017.

\bibitem{safarani2021towards}
S.~Safarani, A.~Nix, K.~F. Willeke, S.~A. Cadena, K.~Restivo, G.~Denfield,
  A.~S. Tolias, and F.~H. Sinz.
\newblock Towards robust vision by multi-task learning on monkey visual cortex.
\newblock In {\em Advances in neural information processing systems (NeurIPS)},
  2021.

\bibitem{Schrimpf2018BrainScoreWA}
M.~Schrimpf, J.~Kubilius, H.~Hong, N.~J. Majaj, R.~Rajalingham, E.~B. Issa,
  K.~Kar, P.~Bashivan, J.~Prescott-Roy, K.~Schmidt, D.~Yamins, and J.~J.
  DiCarlo.
\newblock Brain-score: Which artificial neural network for object recognition
  is most brain-like?
\newblock {\em bioRxiv}, 2018.

\bibitem{Oord2017NeuralD}
A.~van~den Oord, O.~Vinyals, and K.~Kavukcuoglu.
\newblock Neural discrete representation learning.
\newblock In {\em Advances in Neural Information Processing Systems (NeurIPS)},
  2017.

\bibitem{vincent2008extracting}
P.~Vincent, H.~Larochelle, Y.~Bengio, and P.-A. Manzagol.
\newblock Extracting and composing robust features with denoising autoencoders.
\newblock In {\em Proceedings of the 25th international conference on Machine
  learning}, pages 1096--1103, 2008.

\bibitem{Vinje2000SparseCA}
W.~E. Vinje and J.~L. Gallant.
\newblock Sparse coding and decorrelation in primary visual cortex during
  natural vision.
\newblock {\em Science}, 287 5456:1273--6, 2000.

\bibitem{Vintch2015ACS}
B.~Vintch, J.~Movshon, and E.~P. Simoncelli.
\newblock A convolutional subunit model for neuronal responses in macaque v1.
\newblock {\em The Journal of Neuroscience}, 35:14829 -- 14841, 2015.

\bibitem{walker2019inception}
E.~Y. Walker, F.~H. Sinz, E.~Cobos, T.~Muhammad, E.~Froudarakis, P.~G. Fahey,
  A.~S. Ecker, J.~Reimer, X.~Pitkow, and A.~S. Tolias.
\newblock Inception loops discover what excites neurons most using deep
  predictive models.
\newblock {\em Nature neuroscience}, 22(12):2060--2065, 2019.

\bibitem{wang2021encoder}
G.~Wang, W.~Li, L.~Zhang, L.~Sun, P.~Chen, L.~Yu, and X.~Ning.
\newblock Encoder-x: Solving unknown coefficients automatically in polynomial
  fitting by using an autoencoder.
\newblock {\em IEEE Transactions on Neural Networks and Learning Systems},
  2021.

\bibitem{wang2018image}
G.~Wang, J.~C. Ye, K.~Mueller, and J.~A. Fessler.
\newblock Image reconstruction is a new frontier of machine learning.
\newblock {\em IEEE transactions on medical imaging}, 37(6):1289--1296, 2018.

\bibitem{Wang2004ImageQA}
Z.~Wang, A.~C. Bovik, H.~R. Sheikh, and E.~P. Simoncelli.
\newblock Image quality assessment: from error visibility to structural
  similarity.
\newblock {\em IEEE Transactions on Image Processing}, 13:600--612, 2004.

\bibitem{Weliky2003CodingON}
M.~Weliky, J.~Fiser, R.~H. Hunt, and D.~N. Wagner.
\newblock Coding of natural scenes in primary visual cortex.
\newblock {\em Neuron}, 37:703--718, 2003.

\bibitem{Willmore2008TheBW}
B.~Willmore, R.~Prenger, M.~C. Wu, and J.~Gallant.
\newblock The berkeley wavelet transform: A biologically inspired orthogonal
  wavelet transform.
\newblock {\em Neural Computation}, 20:1537--1564, 2008.

\bibitem{Yamins2016UsingGD}
D.~Yamins and J.~DiCarlo.
\newblock Using goal-driven deep learning models to understand sensory cortex.
\newblock {\em Nature Neuroscience}, 19:356--365, 2016.

\bibitem{Yoshida2020NaturalIA}
T.~Yoshida and K.~Ohki.
\newblock Natural images are reliably represented by sparse and variable
  populations of neurons in visual cortex.
\newblock {\em Nature Communications}, 11, 2020.

\bibitem{Zhang2021VisualAI}
J.~Zhang, X.~Zhu, S.~Wang, H.~Esteky, Y.~Tian, R.~Desimone, and H.~Zhou.
\newblock Visual attention in the fovea and the periphery during visual search.
\newblock {\em bioRxiv}, 2021.

\bibitem{Zhuang2017DeepLP}
C.~Zhuang, Y.~Wang, D.~Yamins, and X.~Hu.
\newblock Deep learning predicts correlation between a functional signature of
  higher visual areas and sparse firing of neurons.
\newblock {\em Frontiers in Computational Neuroscience}, 11, 2017.

\bibitem{Zhuang2021UnsupervisedNN}
C.~Zhuang, S.~Yan, A.~Nayebi, M.~Schrimpf, M.~C. Frank, J.~DiCarlo, and
  D.~Yamins.
\newblock Unsupervised neural network models of the ventral visual stream.
\newblock {\em Proceedings of the National Academy of Sciences of the United
  States of America}, 118, 2021.

\end{thebibliography}

}
\end{document}